\documentclass{article}

\usepackage{arxiv}

\usepackage[utf8]{inputenc} 
\usepackage[T1]{fontenc}    
\usepackage{hyperref}       
\usepackage{url}            
\usepackage{booktabs}       
\usepackage{amsfonts}       
\usepackage{nicefrac}       
\usepackage{microtype}      
\usepackage{lipsum}		
\usepackage{graphicx}
\usepackage[numbers]{natbib}
\usepackage{doi}
\usepackage{amsmath}
\usepackage{cleveref}
\usepackage{todonotes}
\usepackage{subcaption}
\usepackage{lscape}
\usepackage{xcolor}
\usepackage{ulem}

\usepackage{makecell}
\usepackage{afterpage}
\usepackage{siunitx}

\usepackage{easyReview}

\setreviewson

\title{Finite Basis Kolmogorov-Arnold Networks: Domain Decomposition for Data-Driven and Physics-Informed Problems}

\date{} 					

\author{
        \href{https://orcid.org/0000-0002-6411-6198}{\includegraphics[scale=0.06]{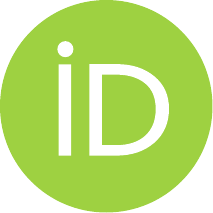}\hspace{1mm}Amanda A. Howard}\\
	Pacific Northwest National Laboratory\\
	Richland, WA 99354 USA \\
	\texttt{amanda.howard@pnnl.gov} \\
 \And
\href{https://orcid.org/0009-0001-5361-3105}{\includegraphics[scale=0.06]{orcid.pdf}\hspace{1mm}Bruno Jacob}\\
 	Pacific Northwest National Laboratory\\
	Richland, WA 99354  USA \\
    \And
 \href{https://orcid.org/0009-0002-1337-0089}{\includegraphics[scale=0.06]{orcid.pdf}\hspace{1mm}Sarah Helfert}\\
             University of North Carolina, Charlotte\\
	Charlotte, NC USA \\
 	Pacific Northwest National Laboratory\\
	Richland, WA 99354  USA \\
    \And
         \href{https://orcid.org/0000-0003-1578-8104}{\includegraphics[scale=0.06]{orcid.pdf}\hspace{1mm}Alexander Heinlein}\\
	 Delft University of Technology\\
	 Delft Institute of Applied Mathematics \\
	 2628 CD Delft, The Netherlands \\
	 \texttt{a.heinlein@tudelft.nl}
    \And
        \href{https://orcid.org/0000-0002-9928-5637}{\includegraphics[scale=0.06]{orcid.pdf}\hspace{1mm}Panos Stinis} \\
	Pacific Northwest National Laboratory\\
	Richland, WA 99354 USA \\
 	University of Washington, Applied Mathematics\\
	Seattle, WA USA\\
 	Brown University, Applied Mathematics\\
	  Providence, RI, 02912 USA \\
        \texttt{panagiotis.stinis@pnnl.gov}
}

\renewcommand{\shorttitle}{Finite Basis Kolmogorov-Arnold Networks}

\hypersetup{
}

\begin{document}
\maketitle

\begin{abstract}
    Kolmogorov-Arnold networks (KANs) have attracted attention recently as an alternative to multilayer perceptrons (MLPs) for scientific machine learning. However, KANs can be expensive to train, even for relatively small networks. Inspired by finite basis physics-informed neural networks (FBPINNs), in this work, we develop a domain decomposition-based architecture for KANs that allows for several small KANs to be trained in parallel to give accurate solutions for multiscale problems. We denote this new approach as finite basis KANs (FBKANs) and show that it yields accurate results for noisy data as well as physics-informed training.
\end{abstract}

\keywords{Kolmogorov-Arnold networks \and Physics-informed neural networks \and Domain decomposition \and Network architecture \and Multiscale problems \and Finite basis PINNs}

\begin{figure}[h]
    \centering
    \includegraphics[width=\textwidth]{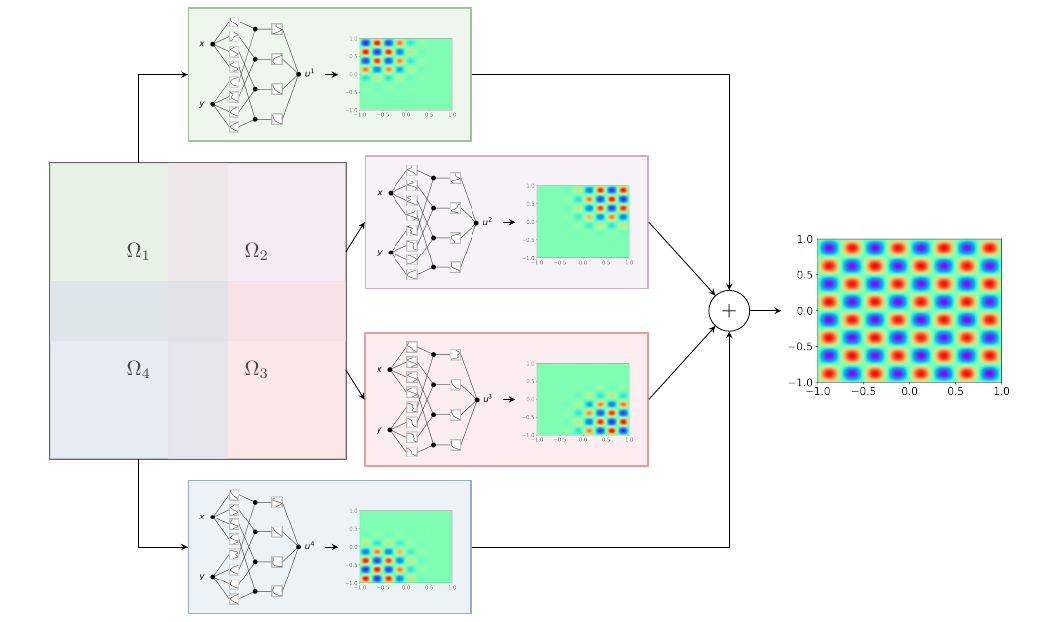}
    \caption{Graphical abstract: The computational domain is decomposed into overlapping subdomains, and an individual KAN model is defined on each subdomain. Then, the global model output is obtained by scaling the outputs of the local KAN models with partition of unity functions and summing up the local contributions.}
    \label{fig:graph_abs}
\end{figure}

\section{Introduction}
Scientific machine learning, predominantly developed using the theory of multilayer perceptrons (MLPs), has  garnered a great deal of attention~\cite{karniadakis2021physics, baker2019workshop, carter2023advanced} over recent years. An alternative model architecture, which is based on the Kolmogorov-Arnold representation theorem~\cite{kolmogorov1957representations}, has recently been introduced in~\cite{liu2024kan}. KANs offer advantages over MLPs in some contexts, such as for continual learning and noisy data. While MLPs use trainable weights and biases on edges with fixed activation functions, KANs employ trainable activation functions represented by splines. This switch may be beneficial because splines are easy to adjust locally and adapt to different resolutions. KANs have also been shown to provide increased interpretability over an MLP. Consequently, it is essential to provide a thorough analysis of the performance and applicability of KANs. An active area of research involves exploring how to utilize a KAN effectively and discerning in which cases KANs might be preferred over classical MLPs. In this paper, we develop an architecture for KANs based on overlapping domain decompositions, which allows for more accurate solutions for complex, and particularly multiscale, problems.

Since the publication of~\cite{liu2024kan}, numerous variations of KANs have been released, including fractional KANs~\cite{aghaei2025fkan}, deep operator KANs~\cite{abueidda2025deepokan}, physics-informed KANs (PI-KANs)\cite{shukla2024comprehensive}, KAN-informed neural networks (KINNs)\cite{wang2025kolmogorov}, temporal KANs~\cite{genet2024tkan}, graph KANs~\cite{kiamari2024gkan, decarlo2024kolmogorovarnoldgraphneuralnetworks, bresson2024kagnnskolmogorovarnoldnetworksmeet}, Chebyshev KANs (cKANs)~\cite{ss2024chebyshev}, convolutional KANs~\cite{Bodner2024}, ReLU-KANs~\cite{qiu2024relu}, among others. KANs have been applied to satellite image classification~\cite{cheon2024kolmogorov}, abnormality detection~\cite{huang2024abnormality}, and computer vision~\cite{azam2024suitability}. Although these results are promising, many open questions remain about how KANs can be best utilized and optimized. Many of the recent variations employ alternative functions to the B-spline parametrization, while others extend KANs through well-developed methods from the field of machine learning, such as physics-informed neural networks (PINNs)~\cite{raissi2019physics}, deep operator networks (DeepONets)~\cite{lu2019deeponet}, and recurrent neural networks. Because KANs still follow the general network architecture of an MLP, aside from the switch between the edges and neurons, it is often straightforward to apply such techniques to KANs.

PINNs~\cite{raissi2019physics} have demonstrated success for a wide range of problems, including fluid dynamics~\cite{cai2021physics, jin2021nsfnets, raissi2020hidden, mahmoudabadbozchelou2022nn, almajid2022prediction}, energy systems~\cite{chen2023physics, misyris2020physics, huang2022applications, moya2023dae, bento2023physics}, and heat transfer~\cite{cai2021physics}. PINNs allow for the efficient and accurate solution of physical problems where robust data is lacking. This is accomplished by incorporating the governing differential equations of a system into the loss function of a neural network, which ensures that the resulting solution will satisfy the physical laws of the system. Although PINNs can successfully train with little or no data in many cases, there are notable cases in which challenges arise. For instance, PINNs can get stuck at the fixed points of dynamical systems during training~\cite{rohrhofer2023ds}. PINNs are also difficult to train for multiscale problems, due to the so-called spectral bias~\cite{rahaman_spectral_2019,wang2022and}, or when the problem domain is large.
Domain decomposition techniques can improve training in such cases~\cite{jagtap2020extended, penwarden2023unified, mattey2022novel, wight2020solving, moseley2023finite}; for a general overview of domain decomposition approaches in scientific machine learning, see also~\cite{heinlein_combining_2021, klawonn_machine_2024}. Rather than training the network over the entire domain, there are many approaches for breaking up the domain into subdomains and training networks on each subdomain.  A popular choice for domain decomposition applied to PINNs is extended PINNs (XPINNs) introduced in~\cite{jagtap2020extended}, with a parallel version presented in~\cite{shukla2021parallel}, and extended in~\cite{penwarden2023unified,hu2022extended}. XPINNs employ a separate subnetwork for each subdomain, which are stitched together across boundaries and trained in parallel. Another class of domain decomposition techniques for PINNs consists of time marching methods, where, in general, a network is initially trained for a small subdomain, and then the subdomain is slowly expanded until it contains the entire domain. Methods in this class include backward-compatible PINNs, which consist of a single network shared by all subdomains~\cite{mattey2022novel}. Also similar is the time marching approach in~\cite{wight2020solving}. Finally, a particularly successful technique is finite basis PINNs (FBPINNs)~\cite{moseley2023finite,dolean_finite_2024, dolean_multilevel_2024, heinlein2024multifidelity}, which use partition of unity functions to weight the output of neural networks in each domain; these have also been extended to randomized neural networks~\cite{shang2025overlapping,anderson_elm-fbpinn_2024}. In comparison to other techniques, FBPINNs offer simplicity because the boundary conditions between adjacent domains do not have to be enforced via additional penalty terms in the loss function. This work explores whether a similar technique can be used to enhance the training of a KAN for challenging cases.

In addition to domain decomposition, many methods have been developed to improve the training of PINNs, including adaptive weighting schemes~\cite{mcclenny2020self, wang2022and}, residual-based attention~\cite{anagnostopoulos2024residual}, adaptive residual point selection~\cite{wu2023comprehensive, mao2023physics, hou2023enhancing, lau2024pinnacle, nabian2021efficient, gao2023failure, visser_pacmann_2024}, causality techniques~\cite{wang2022causality}, multifidelity PINNs~\cite{penwarden2022multifidelity, meng2019multifidelity}, and hierarchical methods to learn more accurate solutions progressively~\cite{howard2023continual, howard2023stacked, wang2024multi, ainsworth2021galerkin, ainsworth2022galerkin, trask2022hierarchical, aldirany2024multi}. A substantial advantage of the methodology presented in this work is that it can easily be combined with the aforementioned techniques to improve the training further.

This paper is structured as follows: We first introduce finite basis KANs (FBKANs) in~\cref{sec:methods}. In~\cref{sec:data-results,sec:physics-results}, we highlight some of the features of FBKANs applied to data-driven and physics-informed problems, respectively. We show that FBKANs can increase accuracy over KANs. One important feature of FBKANs is that the finite basis architecture serves as a wrapper around a KAN architecture. While we have chosen to focus on KANs as described in~\cite{liu2024kan}, most available extensions of KANs could be considered instead to increase accuracy or robustness further.

\section{Methodology}\label{sec:methods}
\subsection{Kolmogorov-Arnold Networks} \label{sec:kans}
In~\cite{liu2024kan}, the authors proposed approximating a multivariate function $f(\mathbf{x})$ by a model of the form
\begin{equation}
    f(\mathbf{x}) \approx \sum_{i_{n_l-1}=1}^{m_{n_l-1}}\varphi_{n_l-1, i_{n_l}, i_{n_l-1}}
    \left(\sum_{i_{n_l-2}=1}^{m_{n_l-2}} \ldots
    \left(\sum_{i_{2}=1}^{m_{2}}\varphi_{2, i_3, i_2}
    \left(\sum_{i_{1}=1}^{m_{1}}\varphi_{1, i_2, i_1}
    \left(\sum_{i_{0}=1}^{m_{0}}\varphi_{0, i_1, i_0}(x_{i_0})
    \right)
    \right)
    \right) \ldots
    \right), \label{eq:KAN}
\end{equation}
which they denote as a Kolmogorov-Arnold network. Here, $n_l$ is the number of layers in the KAN, $\{m_j\}_{j=0}^{n_l}$ is the number of nodes per layer,  $\varphi_{i,j, k}$ are the univariate activation functions, and $\mathbf{x} = (x_1,\ldots,x_{m_0})$. We denote the right-hand side of~\cref{eq:KAN} as {$\mathcal{K}(\mathbf{x})$.} The activation functions $\varphi_{k-1, i_{k}, i_{k-1}}$ in the $k$th layer are polynomials of degree $k$ on a grid with $g$ grid points. They are represented by a weighted combination of a basis function $b(x)$ and a B-spline,
\begin{equation}\label{eq:phi_format}
    \varphi (x) = w_b b(x) + w_s \text{spline}(x),
\end{equation}
where
$$
    b(x) = \frac{x}{1+e^{-x}}
$$
and
$$
    \text{spline}(x) = \sum_i c_i B_i(x).
$$
Here, $B_i(x)$ is a polynomial of degree $k$, and $c_i$, $w_b$, and $w_s$ are trainable parameters.

KANs evaluate the B-splines on a precomputed grid. In one dimension, in a domain $[a, b]$, a grid with $g_1$ intervals has grid points $\{t_0 = a, t_1, t_2, \ldots, t_{g_1}=b\}$; cf.~\cite{liu2024kan}. Grid extension~\cite{liu2024kan} allows for fitting a new, fine-grained spline to a coarse-grained spline, increasing the expressivity of the KAN. The coarse splines are transferred to the fine splines following the procedure described in~\cite{liu2024kan} and \cite{rigas2024adaptive}.

In this work, we do not consider the method for enforcing sparsity of the trainable activation functions as outlined in~\cite{liu2024kan} and instead only consider the mean squared error {(MSE)} in the loss function. Many variations of KANs have been proposed recently. However, in this work, we consider only the formulation as outlined in~\cite{liu2024kan}, although we note that the domain decomposition method presented here could be combined with many variants of KANs.

\subsection{Physics-informed neural networks} \label{sec:pinns}
A physics-informed neural network (PINN) is a neural network (NN) that is trained to approximate the solution of an initial-boundary value problem (IBVP) by optimizing loss terms accounting for initial conditions, boundary conditions, and a residual term using backpropagation to calculate derivatives. In particular, the residual term represents how well the output satisfies the governing differential equation~\cite{raissi2019physics}; for early approaches of combining differential equations with neural networks, we refer to~\cite{dissanayake_neural-network-based_1994, lagaris_artificial_1998}. 
More specifically, we aim at approximating the
solution $f$
of a generic IBVP
\begin{equation}
    \begin{aligned}
        f_t+\mathcal{N}_\mathbf{x} [f] & =0,               &  & x\in \Omega, t\in [0,T],         \\
        f(\mathbf{x},t)                & =g(\mathbf{x},t), &  & x\in \partial\Omega, t\in [0,T], \\
        f(\mathbf{x},0)                & =u(\mathbf{x}),   &  & x\in \Omega,
    \end{aligned}
\end{equation}
over the open domain $\Omega\in\mathbb{R}^d$ with boundary $\partial\Omega.$ Here, $x$ and $t$ represent the spatial and temporal coordinates, respectively, $\mathcal{N}_x$ is the differential operator with respect to $x$, and $g$ and $u$ are given functions representing the boundary and initial conditions, respectively.

The PINN model is then trained to minimize the MSE for the initial conditions, boundary conditions, and residual (physics) sampled at $N_{ic}, N_{bc}, N_{r}$ data points. We denote the sampling points for the initial conditions, boundary conditions, and residual as $\{ x_{ic}^i, u(x_{ic}^i)\}_{i=1}^{N_{ic}}$, $\{(x_{bc}^i,t_{bc}^i), g(x_{bc}^i,t_{bc}^i)\}_{i=1}^{N_{bc}}$, and $\{(x_r^i, t_r^i)\}_{i=1}^{N_r}$, respectively. {Then, we} optimize the following weighted loss function with respect to the trainable parameters $\theta$ of the NN model $f_{\theta}$:
\begin{align*}
    \mathcal{L}(\theta)      & =\lambda_{ic}\mathcal{L}_{ic}(\theta)+\lambda_{bc}\mathcal{L}_{bc}(\theta)+\lambda_{r}\mathcal{L}_{r}(\theta),                                                           \\
    \mathcal{L}_{ic}(\theta) & = \frac{1}{N_{ic}}\sum_{i=1}^{N_{ic}}\left(f_{\theta}(\mathbf{x}_{ic}^i,0)-u(\mathbf{x}_{ic}^i)\right)^2,                                                                \\
    \mathcal{L}_{bc}(\theta) & = \frac{1}{N_{bc}}\sum_{i=1}^{N_{bc}}\left(f_{\theta}(\mathbf{x}_{bc}^i,t_{bc}^i)-g(\mathbf{x}_{bc}^i, t_{bc}^i)\right)^2,                                               \\
    \mathcal{L}_r(\theta)    & = \frac{1}{N_{r}}\sum_{i=1}^{N_r}\left(\frac{\partial}{\partial t} f_{\theta}(\mathbf{x}_r^i, t_r^i)+\mathcal{N}_\mathbf{x}[f_{\theta}(\mathbf{x}_r^i, t_r^i)]\right)^2,
\end{align*}
where $\lambda_{ic}, \lambda_{bc}, \lambda_{r}$ are weights for the different loss terms.
The choice of these weights can have a significant impact on the training process and model performance. While adaptive schemes are available for choosing the weights~\cite{mcclenny2020self, wang2022and, wang2022improved, howard2024conjugate}, in this work, we set the weights manually to approximately scale the terms in the loss function to have equal magnitude. The values of the weights used are given in Appendix~\ref{sec:training_params}.

\subsection{Finite basis physics-informed neural networks} \label{sec:fbpinns}
\begin{figure}[t]
    \centering
    \includegraphics[width=0.6\textwidth]{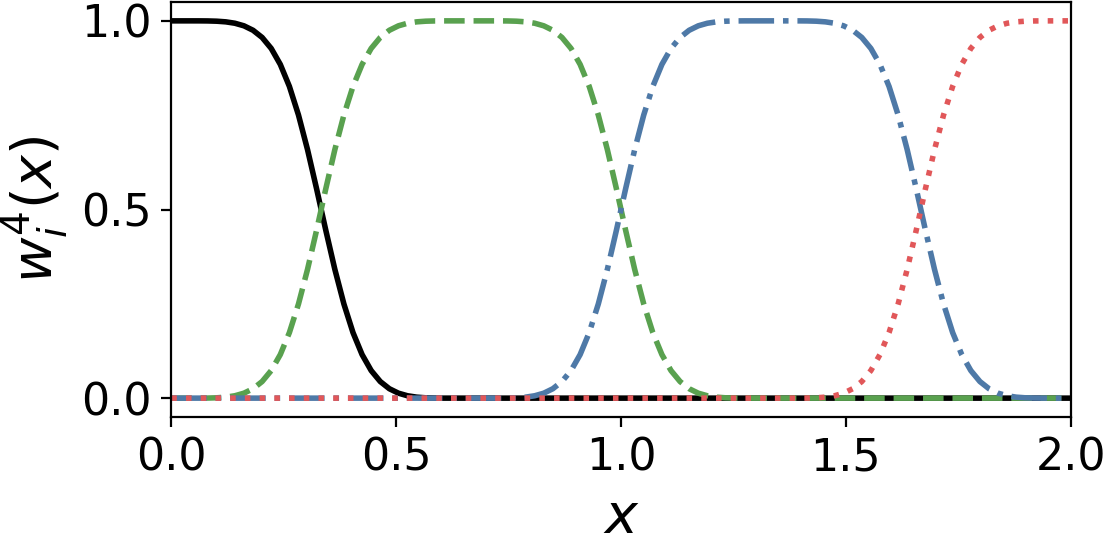}
    \caption{Example of the partition of unity functions on the domain $\Omega = [0, 2]$ with $L=4$ subdomains.}
    \label{fig:POU_1d}
\end{figure}

\begin{figure}[t]
    \centering
    \includegraphics[height=0.35\textwidth]{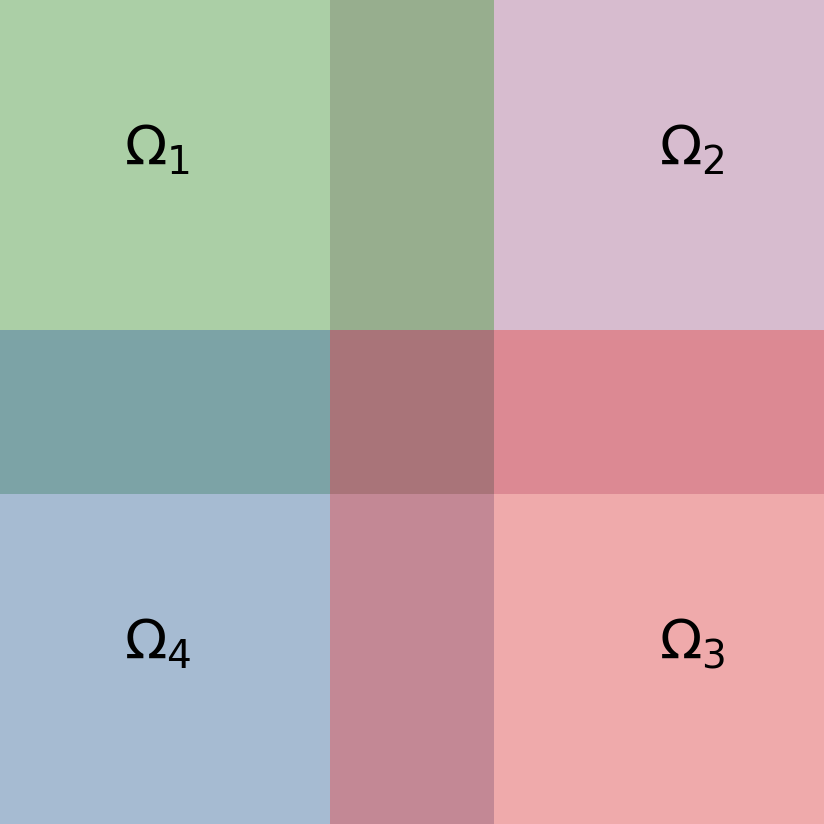}
    \hspace*{0.1\textwidth}
    \includegraphics[height=0.35\textwidth]{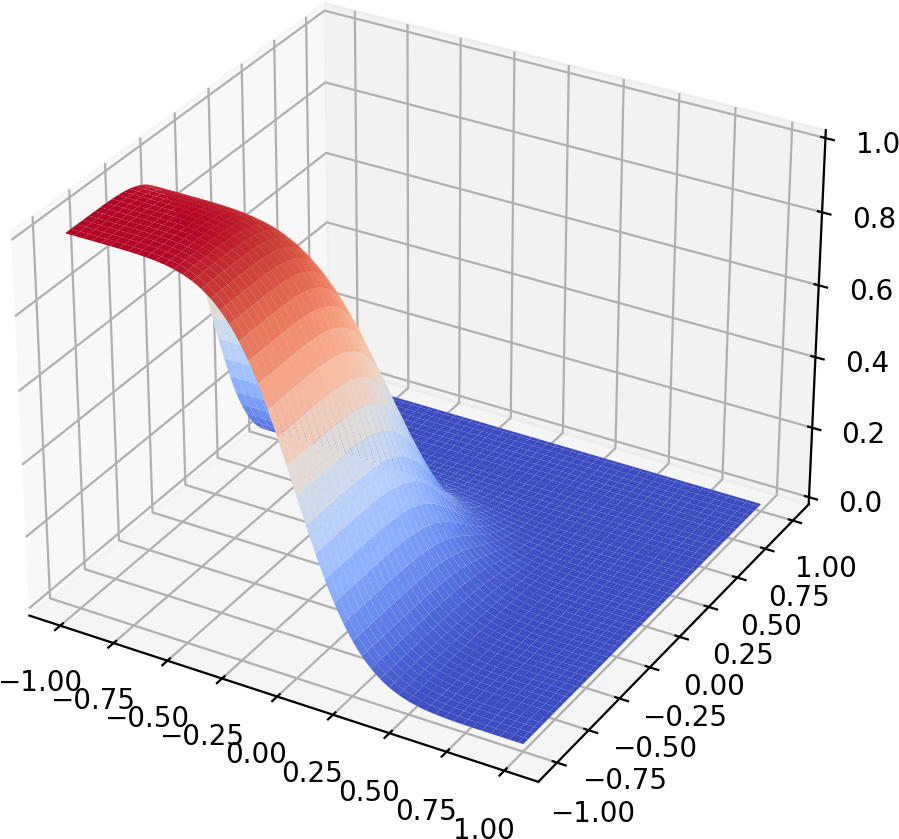}
    \caption{(Left) Example domain decomposition on the domain $\Omega = [-1, 1]\times[-1, 1]$ with $L=4$ subdomains. (Right) One example partition of unity function $\omega_{11}(x, y).$
    }
    \label{fig:POU_2d}
\end{figure}

In finite basis physics-informed neural networks (FBPINNs)~\cite{moseley2023finite, dolean_finite_2024, dolean_multilevel_2024, heinlein2024multifidelity}, we decompose the spatial domain $\Omega$ or the space-time domain $\Omega \times [0,T]$ into overlapping subdomains. Each overlapping subdomain $\Omega_j$ is the interior of the support of a corresponding function $\omega_j$, and all functions $\omega_j$ form a partition of unity. In particular, in the case of $L$ overlapping subdomains, we have
$$
    \Omega = \bigcup_{j=1}^L \Omega_j,
    \quad
    {\rm supp}(\omega_j) = \overline{\Omega_j},
    \text{ and}
    \quad
    \sum_{j=1}^{L} \omega_j \equiv 1 \text{ in } \Omega.
$$
In one dimension, a uniform overlapping domain decomposition with overlap ratio $\delta>1$ is given by subdomains
$$
    \Omega_j = \left(\frac{(j-1)l -\delta l/2}{L-1}, \frac{(j-1)l +\delta l/2}{L-1} \right),
$$
where
$l = \max(x) - \min(x)$.
Then, there are multiple ways to define the partition of unity functions. Here, we construct them based on the expression
$$
    \omega_j = \frac{\hat\omega_j}{\sum_{j=1}^{L} \hat\omega_j},
$$
where
$$ 
    \hat\omega_j(x) = \begin{cases}
        1                                                        & L = 1, \\
        \left[ 1+\cos\left(\pi(x-\mu_j)/\sigma_j\right)\right]^2 & L>1,
    \end{cases}
$$
with $\mu_j = l(j-1)/(L-1)$ and $\sigma_j = (\delta l/2)/(L-1)$ representing the center and half-width of each subdomain, respectively. An example of the one-dimensional partition of unity functions with $L=4$ is depicted in~\cref{fig:POU_1d}.
In multiple dimensions, $\mathbf{x}\in \mathbb{R}^d$, we then obtain the partition of unity functions as the tensor product of the one-dimensional functions:
$$
    \hat\omega_j(\mathbf{x}) = \begin{cases}
        1                                                                            & L = 1, \\
        \Pi_{i=1}^d \left[ 1+\cos\left(\pi(x_i-\mu_{ij})/\sigma_{ij}\right)\right]^2 & L>1,
    \end{cases} 
$$
with $\mu_{ij}$ and $\sigma_{ij}$ representing the center and half-width of each subdomain $j$ in each dimension $i$. An example of an overlapping domain decomposition and of a single partition of unity function with $d=2$ and $L=4$ are shown in~\cref{fig:POU_2d}.

Then, the FBPINN architecture reads
\begin{equation}
    f_\theta (\mathbf{x}) = \sum_{j=1}^{L} \omega_j (x) f_{j} (\mathbf{x}; \theta^j), \label{eq:FBPINN}
\end{equation}
where $f_{j}(\cdot; \theta^j)$ is the neural network with parameters $\theta^j$ that corresponds to the subdomain $\Omega_j$; it is localized to $\Omega_j$ by multiplication with the partition of unity function $\omega_j$, which is zero outside of $\Omega_j$. The FBPINN model is trained in the same way as the PINN model, i.e., using initial, boundary, and residual loss terms; cf.~\cref{sec:pinns}. Note that, in the original FBPINNs paper~\cite{moseley2023finite}, hard enforcement of initial and boundary conditions is employed, such that only the residual loss term remains. Moreover, in~\cite{dolean_multilevel_2024} the approach has been extended to multilevel domain decompositions, in~\cite{heinlein2024multifidelity} this domain decomposition approach has been applied to long-range time-dependent problems, and in~\cite{shang2025overlapping,anderson_elm-fbpinn_2024} it has been extended to randomized neural networks.

\subsection{Finite basis Kolmogorov--Arnold networks} \label{sec:fbkans}
In finite basis Kolmogorov--Arnold networks (FBKANs), we replace the NN used in FBPINNs with a KAN.
The function approximation in~\cref{eq:KAN} then becomes
\begin{align}
    f(\mathbf{x}) & \approx \sum_{j=1}^L \omega_j(x)
    \left[
    \sum_{i_{n_l-1}=1}^{m_{n_l-1}}\varphi^j_{n_l-1, i_{n_l}, i_{n_l-1}}
    \left(\sum_{i_{n_l-2}=1}^{m_{n_l-2}} \ldots
    \left(\sum_{i_{1}=1}^{m_{1}}\varphi^j_{1, i_2, i_1}
    \left(\sum_{i_{0}=1}^{m_{0}}\varphi^j_{0, i_1, i_0}(x_{i_0})
    \right)
    \right) \ldots
    \right)
    \right] \nonumber                                                                                \\
                  & = \sum_{j=1}^L \omega_j(x) \mathcal{K}_j (\mathbf{x}; \theta^j), \label{eq:FBKAN}
\end{align}

where $\mathcal{K}_j(\mathbf{x}; \theta^j)$ denotes the $j^{th}$ KAN with trainable parameters $\theta^j$
and $\omega_j$ denotes the corresponding finite basis partition of unity function introduced in~\cref{sec:fbpinns}. FBKANs are trained to minimize the {PINN} loss function
\begin{equation}
    \mathcal{L}(\theta) = \lambda_{ic}\mathcal{L}_{ic}(\theta)+\lambda_{bc}\mathcal{L}_{bc}(\theta)+\lambda_{r}\mathcal{L}_{r}(\theta)+\lambda_{data}\mathcal{L}_{data}(\theta), \label{eq:loss_KAN}
\end{equation}
composed of
\begin{align*}
    \mathcal{L}_{ic}(\theta)   & = \frac{1}{N_{ic}}\sum_{i=1}^{N_{ic}}\left(\sum_{j=1}^L \omega_j(\mathbf{x}) \mathcal{K}_j(\mathbf{x}_{ic}^i; \theta^j)-u(\mathbf{x}_{ic}^i)\right)^2,                                                                                                                                \\
    \mathcal{L}_{bc}(\theta)   & = \frac{1}{N_{bc}}\sum_{i=1}^{N_{bc}}\left(\sum_{j=1}^L \omega_j(\mathbf{x}) \mathcal{K}_j(\mathbf{x}_{bc}^i; \theta^j)-g(\mathbf{x}_{bc}^i, t_{bc}^i)\right)^2,                                                                                                                      \\
    \mathcal{L}_r(\theta)      & = \frac{1}{N_{r}}\sum_{i=1}^{N_r}\left(\frac{\partial}{\partial t} \left[\sum_{j=1}^L \omega_j(\mathbf{x}) \mathcal{K}_j(\mathbf{x}_{r}^i; \theta^j)\right]+\mathcal{N}_\mathbf{x} \left[\sum_{j=1}^L \omega_j(\mathbf{x}) \mathcal{K}_j(\mathbf{x}_{r}^i; \theta^j)\right]\right)^2, \\
    \mathcal{L}_{data}(\theta) & = \frac{1}{N_{data}}\sum_{i=1}^{N_{data}}\left(\sum_{j=1}^L \omega_j(\mathbf{x}) \mathcal{K}_j(\mathbf{x}_{data}^i; \theta^j)-f(\mathbf{x}_{data}^i)\right)^2,
\end{align*}
{using the FBKAN model~\cref{eq:FBKAN}.} Here, $\theta = \{\theta^j\}_{j=1}^L$ is the set of trainable parameters, and the loss function $\mathcal{L}$ contains both a term for data-driven training $\mathcal{L}_{data}$ and terms for physics-informed training $\mathcal{L}_{ic}$, $\mathcal{L}_{bc}$, and $\mathcal{L}_{r}$.
In this way, FBKANs are adaptable to given problem characteristics.

For FBKANs, the grid of each local KAN $\mathcal{K}_j$ is defined separately on each subdomain $\Omega^j$. To do so, we densely sample points $x\in \Omega$ and compute the partition of unity function $\omega_j(x)$. We then take the boundaries of each subdomain based on the partition of unity functions as
\begin{align*}
    a^j & = \min_{\substack{x \text{ such that} \\ \omega^j(x) > 10^{-4}}} x, \\
    b^j & = \max_{\substack{x \text{ such that} \\ \omega^j(x) > 10^{-4}}} x.
\end{align*}
Hence, we obtain a custom grid for each subdomain.

We evaluate the model performance using the relative $\ell_2$ error, calculated by
$$
    \frac{||f(x) - \sum_{j=1}^L \omega_j(x)\mathcal{K}_j(x; \theta^j)||_2}{||f(x)||_2},
$$
where $f$ is the ground truth solution.

Our implementation is based on \texttt{JAX}~\cite{jax2018github}, using the \texttt{jaxKAN} package~\cite{Rigas_jaxKAN_A_JAX-based_2024} for the KAN implementation. For simplicity, we employ the same hyperparameters for local KAN models defined on the subdomains, including the width and depth; this reduces the total number of hyperparameters. However, in general, the parameters for each individual KAN $\mathcal{K}_j$ could be chosen differently.

\section{Data-driven results} \label{sec:data-results}
In this section, we consider data-driven problems, which are trained without physics meaning that $\lambda_{ic}=\lambda_{bc}=\lambda_{r}=0$. These results specifically illustrate how FBKANs can improve predictions for noisy data and multiscale oscillations.

\subsection{Data-driven test 1}\label{sec:Test1}
In this section, we consider a toy problem {of fitting the function}
\begin{equation}
    f(x) = \exp[\sin(0.3 \pi x^2)], \label{eq:Case_1}
\end{equation}
on $x \in [0, 8]$. This problem is designed to test both the scaling of FBKANs as we increase the number of subdomains, as well as the impact of noisy data on training FBKANs.

\subsubsection{Scaling of FBKANs} \label{sec:Test1a}
\begin{figure}[t]
    \centering
    \includegraphics[width=\textwidth]{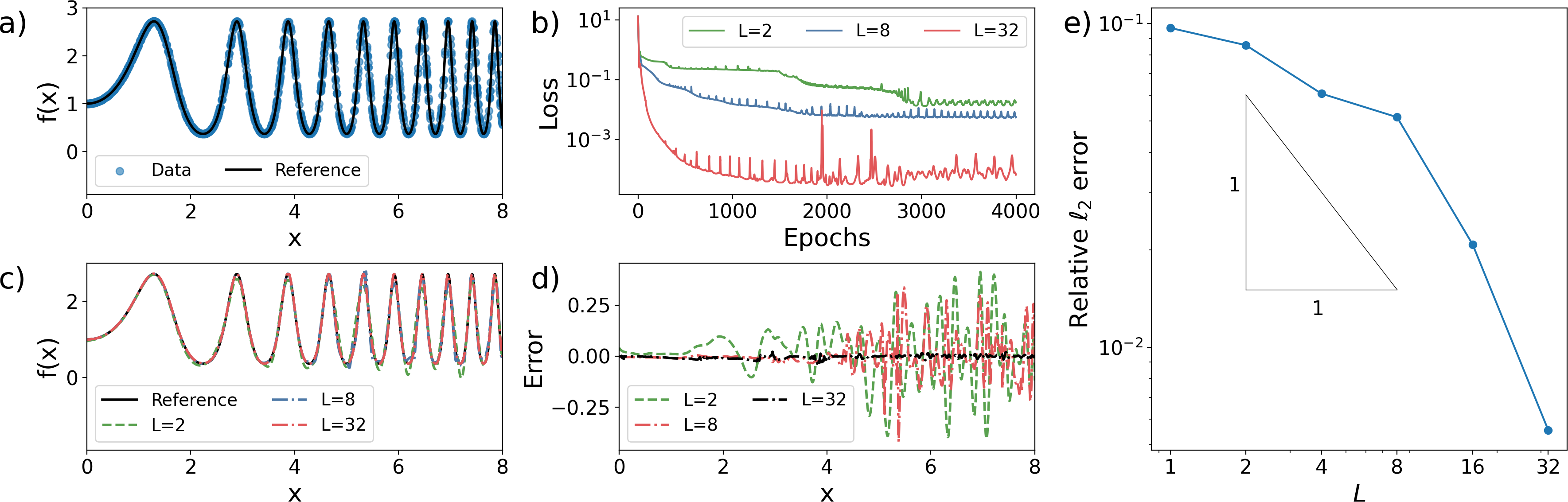}
    \caption{Results with $L=2, 8,$ and $32$ subdomains for~\cref{eq:Case_1}. (a) Training data and plot of exact $f(x)$. (b) Loss curves (\cref{eq:loss_KAN}). (c) Plot of the outputs $f(x)$ and $\sum_{j=1}^L \omega_j(x)\mathcal{K}_j(x)$. (d) Pointwise errors $f(x) - \sum_{j=1}^L \omega_j(x)\mathcal{K}_j(x)$. (e) Scaling results for Test 1 with $L$ subdomains.
    }
    \label{fig:Test1a}
\end{figure}

First, we consider clean data, $N_{data} = 1\,200$, which is sampled uniformly from the interval $[0, 8].$ We present results for training on $L = 1, 2, 4, 8, 16,$ and $32$ subdomains.
Note that the function described in~\cref{eq:Case_1} is highly oscillatory, which makes it difficult to capture its behavior with a small KAN. As shown in~\cref{fig:Test1a}, increasing the number of subdomains from $L=2$ to $L=8$ and $L=32$ significantly decreases the pointwise error in the FBKAN predictions. Doubling the number of subdomains doubles the number of trainable parameters of the FBKAN model, increasing its expressivity. We observe approximately first order convergence of the relative $\ell_2$ error for a large number of subdomains in~\cref{fig:Test1a}.

\subsubsection{FBKANs with noisy data}\label{sec:Test1b}
One strength of KANs over MLPs is their increased accuracy for noisy data~\cite{liu2024kan}. We now test FBKANs on noisy training data using four subdomains ($L=4$). Therefore, we sample \num{600} training points from a uniform distribution in $[0, 8]$ and evaluate $f(x)$. Then, we add Gaussian white noise with zero mean and varying magnitude to $f(x)$, up to an average relative magnitude of $20\,\%$, as the training data set. The method is then tested on \num{1000} evenly spaced points in $[0, 8]$ without noise.

\begin{figure}[t]
    \centering
    \includegraphics[width=\textwidth]{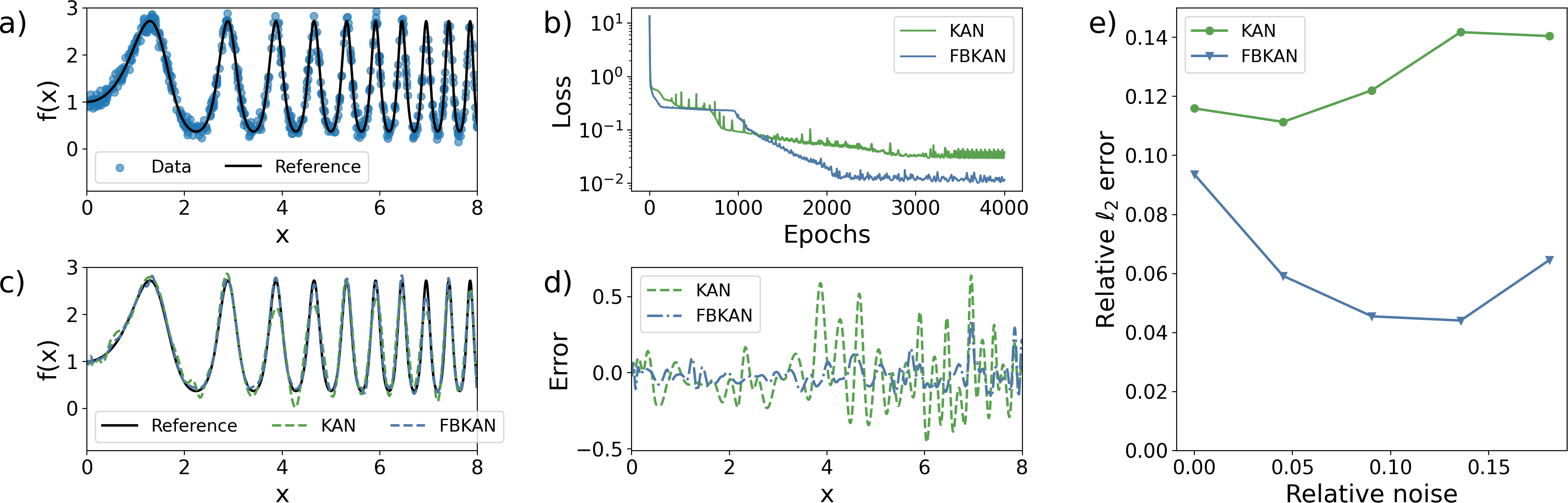}
    \caption{Results for Test 1 with noisy training data; cf.~\cref{eq:Case_1}. (a) Example training data and plot of exact $f(x)$ with $9.6\%$ mean relative noise. (b) Loss curves (\cref{eq:loss_KAN}) for an example training with $9.6\%$ mean relative noise. (c) Plot of the outputs $f(x)$ and $\sum_{j=1}^L \omega_j(x)\mathcal{K}_j(x)$ with $9.6\%$ mean relative noise. (d) Pointwise errors $f(x) - \sum_{j=1}^L \omega_j(x)\mathcal{K}_j(x)$ with $9.6\%$ mean relative noise. (e) Relative $\ell_2$ error of the KANs and FBKANs with respect to the magnitude of the noise added to the training data.}
    \label{fig:Test1b}
\end{figure}

The training set and results are shown in~\cref{fig:Test1b}.
In the case with the largest noise, with relative noise of 18.1\% added to the training set, the KAN yields a relative $\ell_2$ error of 0.1404, whereas the FBKAN yields a relative error of 0.0646. For all noise levels tested, the FBKAN consistently has a lower relative error than the plain KAN in~\cref{fig:Test1b}, and the predictions are robust to noisy data.

\subsection{Data-driven test 2}\label{sec:Test2}
{Next, we consider fitting the function}
\begin{equation}
    f(x, y) = \sin(6 \pi x^2 ) \sin(8 \pi y^2),
\end{equation}
for $(x, y) \in [0, 1]\times [0, 1].$ This data-driven example exhibits fine-scale oscillations. We test two cases: (1) a KAN and FBKAN with a fixed grid, denoted by KAN-1 / FBKAN-1 and (2) a KAN and FBKAN with grid extension, denoted by KAN-2 / FBKAN-2. The FBKAN has $L=4$ subdomains. The training set is composed of \num{10000} points randomly sampled from a uniform distribution.

\begin{figure}[t]
    \centering
    \includegraphics[width=0.55\textwidth]{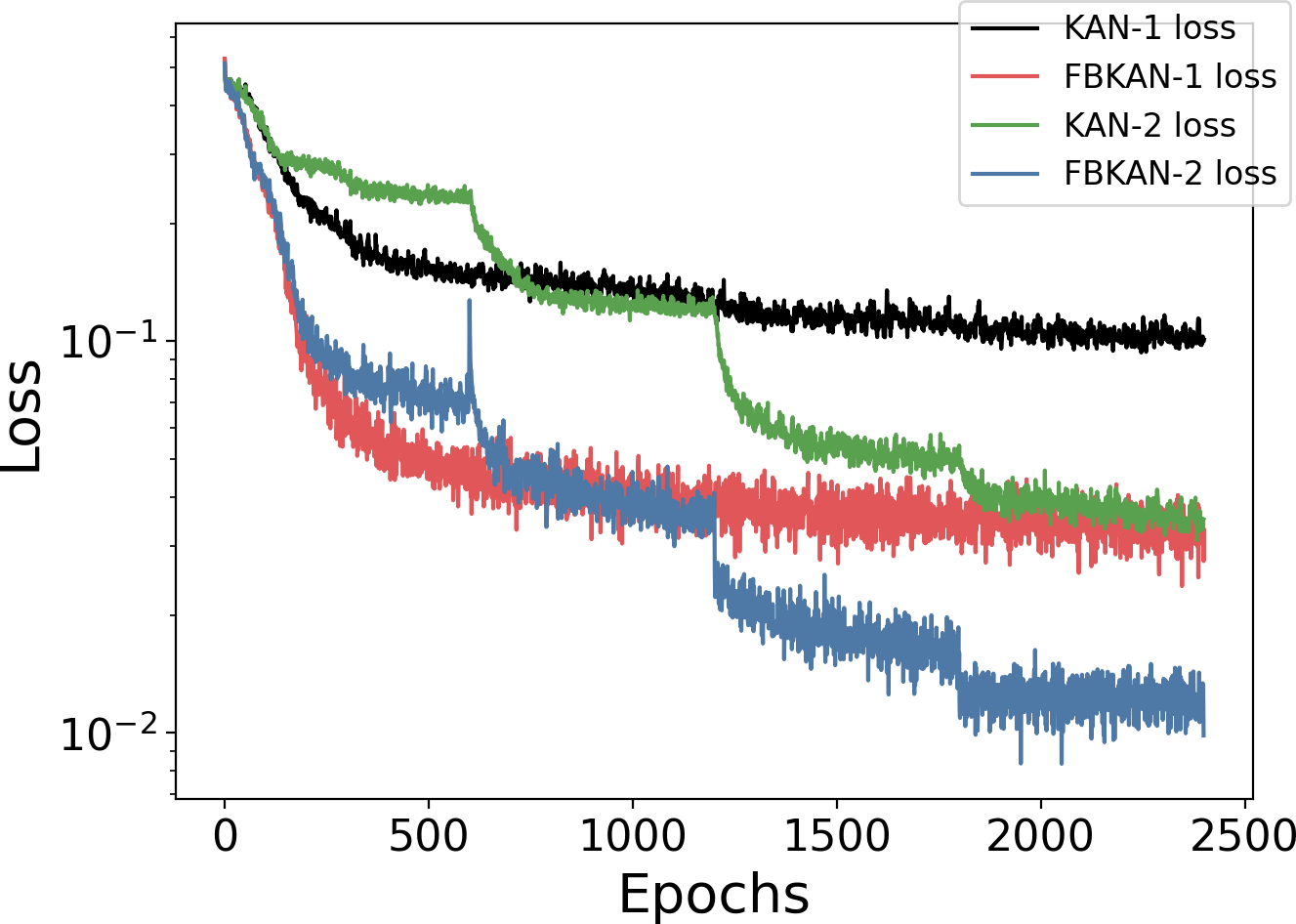}
    \caption{Training loss curves for the data-driven test 2.}
    \label{fig:Test2a}
\end{figure}

\begin{figure}[th]
    \centering
    \includegraphics[width=0.9\textwidth]{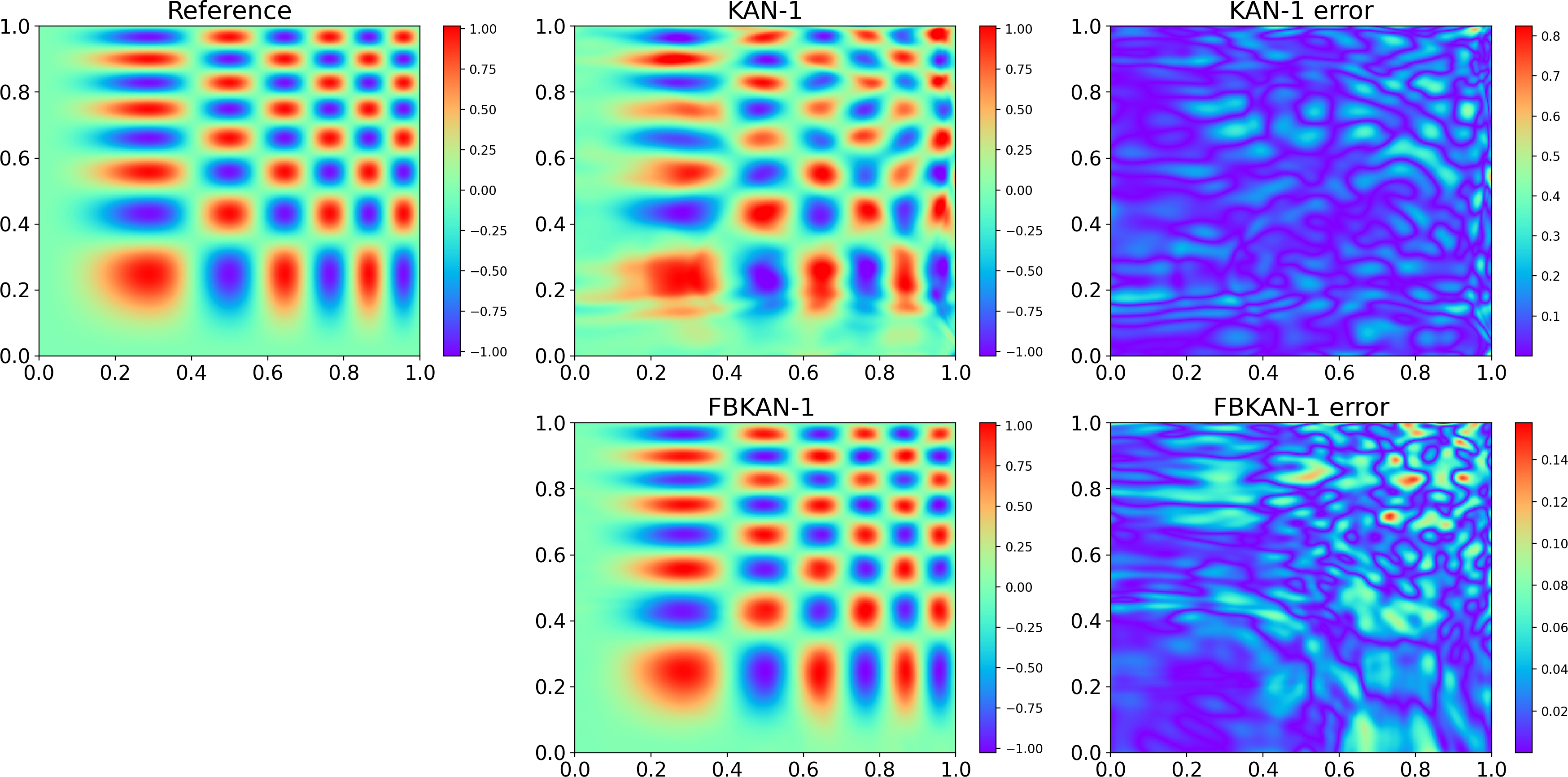}
    \caption{Reference solution (left), predicted solutions (middle) and relative errors (right) for KAN-1 and FBKAN-1 for {data-driven test 2}. Note that the error plots have different color scales.}
    \label{fig:Test2b}
\end{figure}

\begin{figure}[th]
    \centering
    \includegraphics[width=0.9\textwidth]{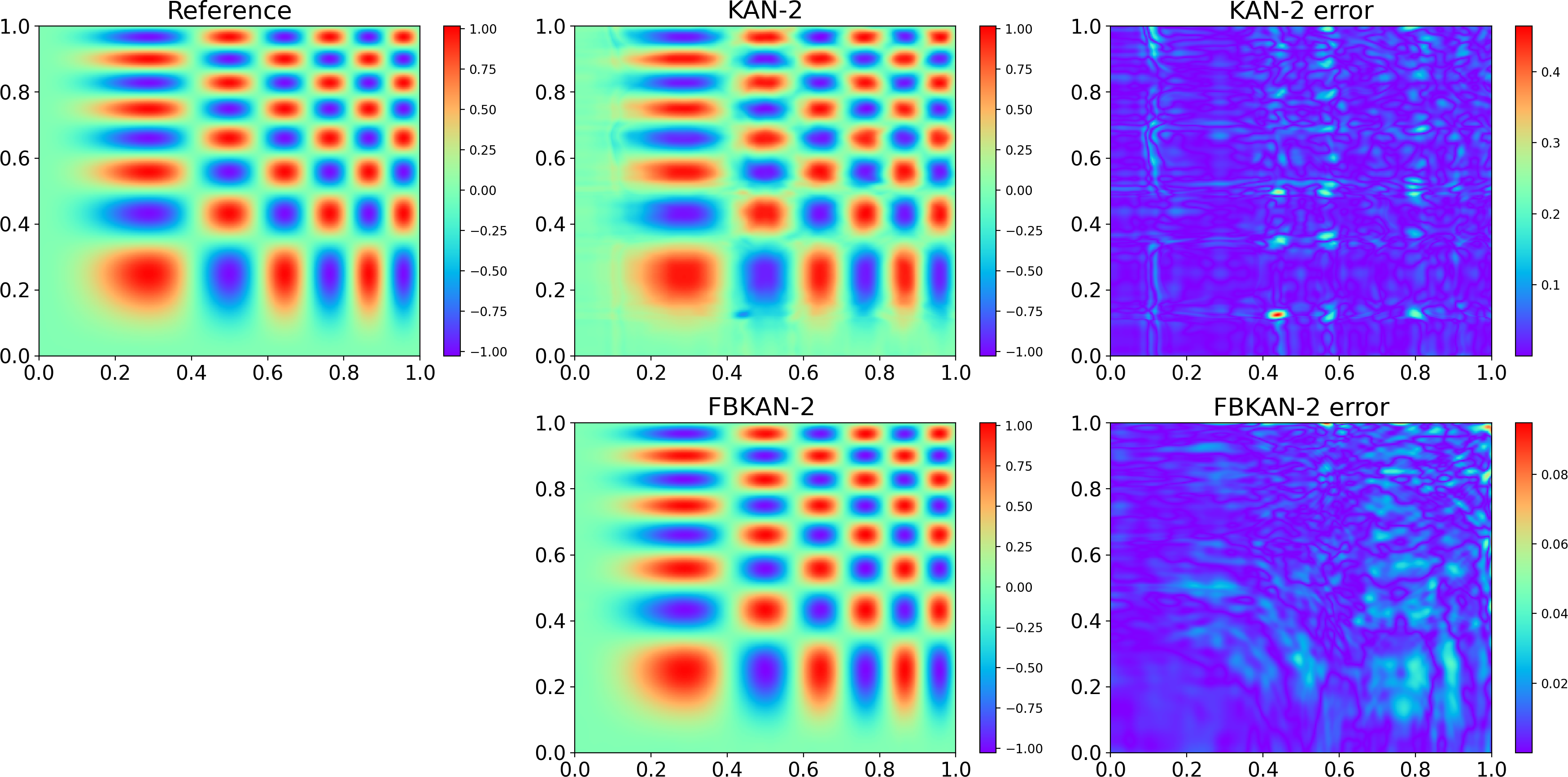}
    \caption{Reference solution (left), predicted solutions (middle) and relative errors (right) for KAN-2 and FBKAN-2 for {data-driven test 2}. Note that the error plots have different color scales.}
    \label{fig:Test2c}
\end{figure}

In both cases, we begin training with $g=5$ grid points. In the grid extension case, the grid increases every \num{600} iterations as listed in~\cref{tab:params_Test2}, and the learning rate drops by $20\,\%$ each time the grid is updated. As can be observed in~\cref{fig:Test2a}, the final training loss for FBKAN-1 with a fixed grid is approximately the same as the training loss for KAN-2 with the grid extension approach, even though each FBKAN network has six times fewer grid points.
This is also reflected in the relative errors reported in~\cref{tab:test2}. Comparing the KAN-1 and FBKAN-1 models in~\cref{fig:Test2b}, KAN-1 struggles to capture the fine-scale features of the data accurately and has a larger relative error. As can be seen in~\cref{fig:Test2c}, KAN-2 and FBKAN-2 both outperform their counterparts with a static grid, but FBKAN-2 is better able to capture the data than KAN-2.

\begin{table}[th]
    \centering
    \begin{tabular}{l l r}
        \toprule
        name    & grid type      & relative error        \\
        \midrule
        KAN-1   & fixed grid     & $2.36 \times 10^{-1}$ \\
        FBKAN-1 & fixed grid     & $7.43 \times 10^{-2}$ \\
        KAN-2   & grid extension & $8.10 \times 10^{-2}$ \\
        FBKAN-2 & grid extension & $2.27 \times 10^{-2}$ \\
        \bottomrule
    \end{tabular}
    \caption{Relative $\ell_2$ errors for {data-driven test 2}.}
    \label{tab:test2}
\end{table}

\section{Physics-informed results} \label{sec:physics-results}
\subsection{Physics-informed test 1}\label{sec:Test3}
As the first physics-informed example, we consider the following one-dimensional problem with multiscale features:
\begin{align*}
    \frac{df}{dx} & = 4\cos(4x) + 40\cos(40 x), \\
    f(0)          & = 0,
\end{align*}
{for} $x \in [-4, 4]$. The analytical solution reads
\begin{equation}
    f(x) = \sin(4x) + \sin(40 x).
\end{equation}

We sample \num{400} collocation points from a uniform distribution ($N_r = 400$) at every training iteration. We start with a grid with $g=5$ and extend it by $5$ every \num{1000} iterations. The KAN reaches a relative $\ell_2$ error of 0.2407 with $g_{final}=20$, whereas the FBKAN for four subdomains ($L=4$) reaches a relative $\ell_2$ error of 0.0898 for $g_{final}=20$. For eight subdomains ($L=8$), the FBKAN reaches a relative error of 0.0369 with $g_{final}=15$ and 0.0662 with $g_{final}=10$.

\begin{figure}[th]
    \centering
    \includegraphics[width=0.9\textwidth]{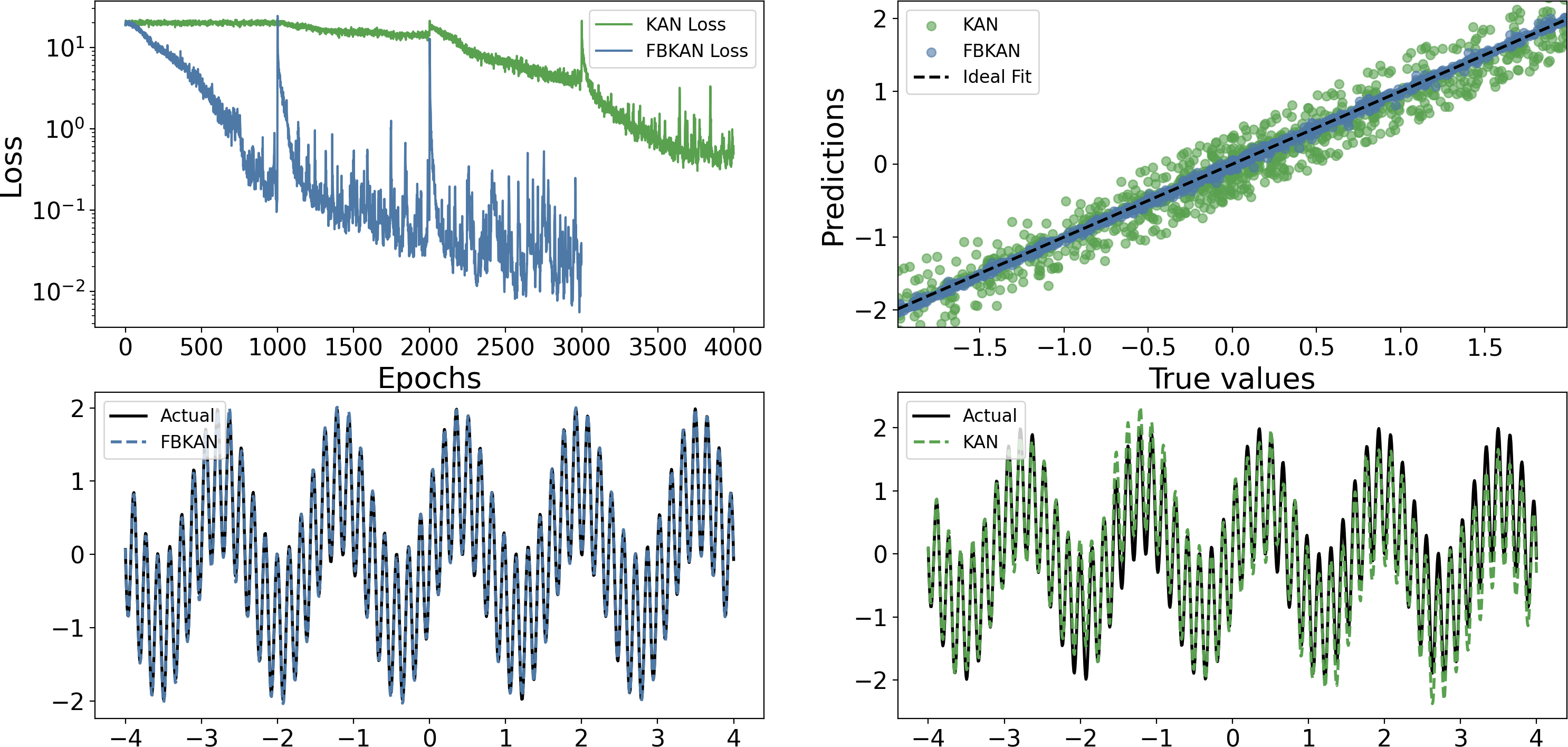}
    \caption{Results for {physics-informed test 1} with $L=8$. The FBKAN clearly outperforms the KAN, reaching a lower loss with fewer grid points.}
    \label{fig:Test3}
\end{figure}

\subsection{Physics-informed test 2}\label{sec:Test4}
Next, we test the Helmholtz equation
\begin{align*}
    \frac{\partial^2 f}{\partial y^2} + \frac{\partial^2 f}{\partial x^2} + k_h^2 f - q(x, y) & = 0, \ (x, y) \in [-1, 1]\times[-1, 1], \\
    f(-1, y) = f(1, y)                                                                        & = 0, \  y \in [-1, 1],                  \\
    f(x, -1) = f(x, 1)                                                                        & = 0, \  x \in [-1, 1],
\end{align*}
with
$$
    q(x, y)
    =
    -(a_1 \pi)^2 \sin(a_1 \pi x) \sin(a_2 \pi y)-(a_2 \pi)^2 \sin(a_1 \pi x) \sin(a_2 \pi y) + k_h^2 \sin(a_1 \pi x) \sin(a_2 \pi y)
$$

which is obtained by using the solution
\begin{align*}
    f(x,y) = \sin(a_1 \pi x) \sin(a_2 \pi y).
\end{align*}

In our tests, we vary $a_1$ and $a_2$.

For each choice of $(a_1, a_2)$ we consider three training cases. In the first case, we use higher-order splines with $k=5$ and a fixed grid with $g = 5$ (denoted by KAN-1 / FBKAN-1). In the second case, we take $k=3$ and use grid extension to increase $g$ by 5 every \num{600} iterations, starting with $g=5$ (denoted by KAN-2 / FBKAN-2). We use one hidden layer with width 10 for both cases. Additionally, we test a third case, denoted by KAN-3 / FBKAN-3, where we consider the same hyperparameters as KAN-1 / FBKAN-1 but with width 5 in the hidden layer. The relative $\ell_2$ errors are reported in~\cref{tab:error-pi2}.

\begin{figure}[th]
    \centering
    \includegraphics[width=0.9\textwidth]{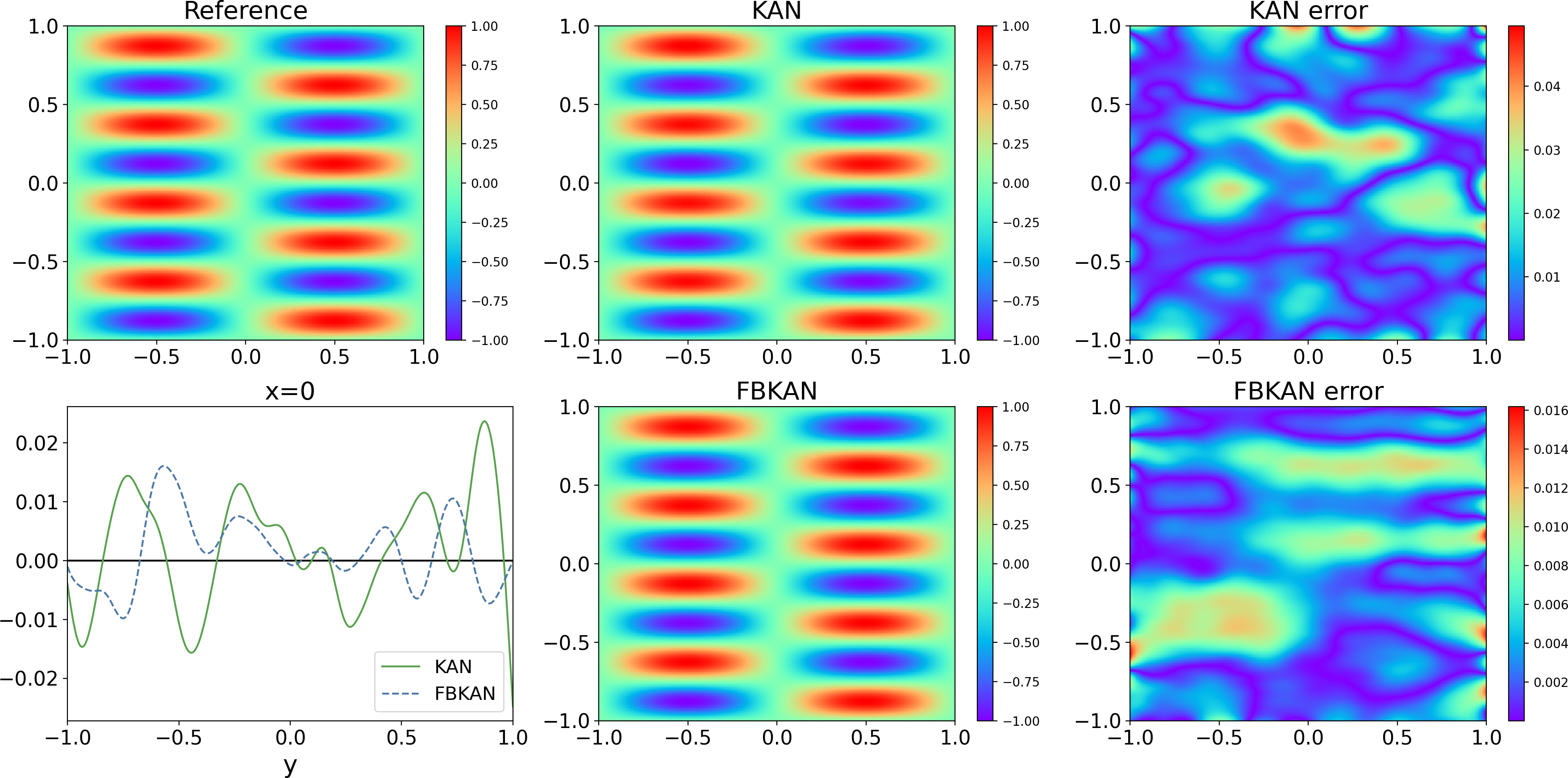}
    \caption{Results for {physics-informed test 2} with KAN-1 and FBKAN-1, and $a_1 = 1, a_2 = 4$, $L=4$. The solution along the line $x=0$ is given in the bottom left subfigure. Note that the error plots have different color scales.}
    \label{fig:Test4a}
\end{figure}

\begin{figure}[th]
    \centering
    \includegraphics[width=0.9\textwidth]{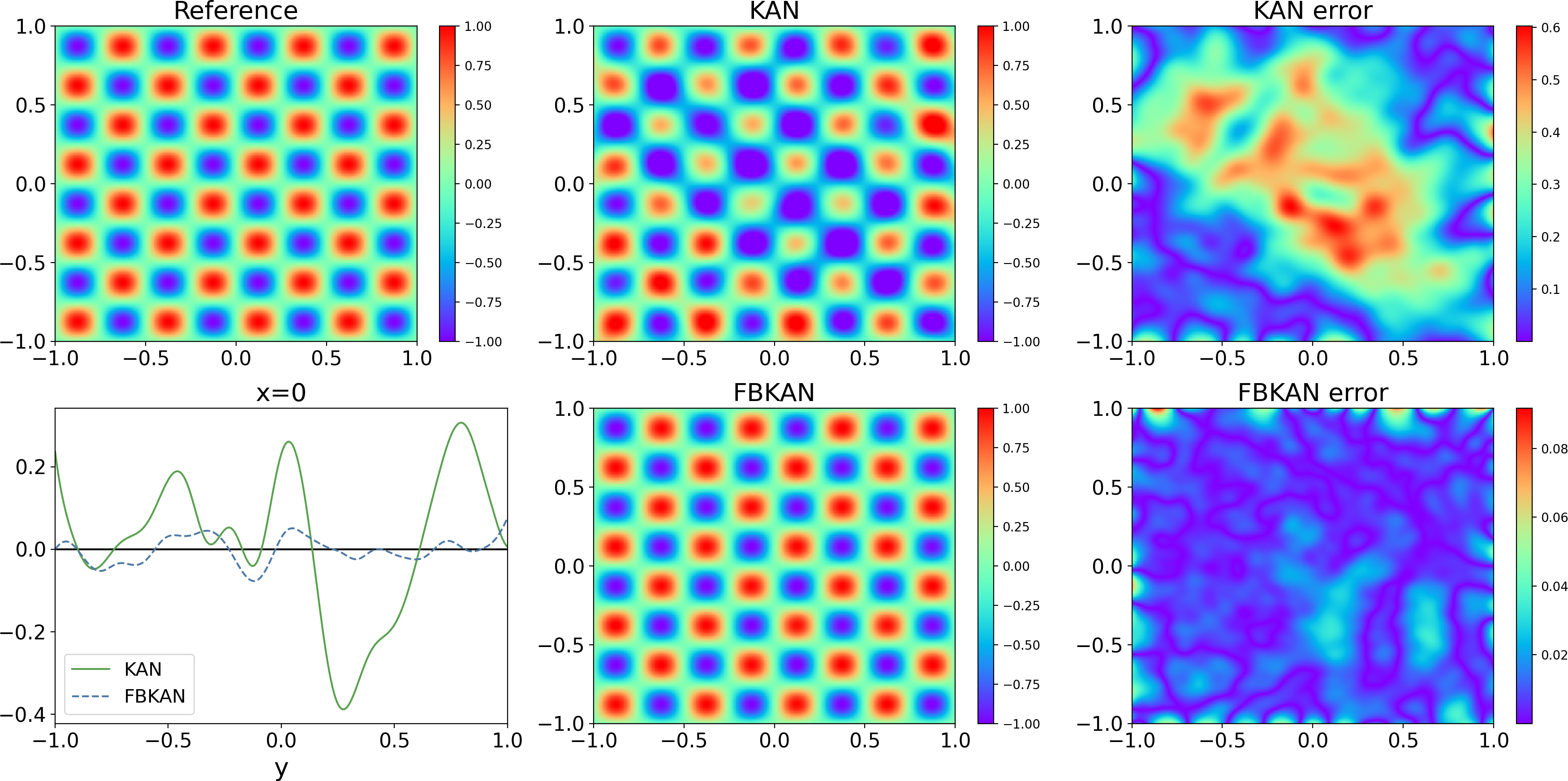}
    \caption{Results for {physics-informed test 2} with KAN-1 and FBKAN-1, and $a_1 = 4, a_2 = 4$, $L=4$. The solution along the line $x=0$ is given in the bottom left subfigure. Note that the error plots have different color scales.}
    \label{fig:Test4b}
\end{figure}

\begin{figure}[th]
    \centering
    \includegraphics[width=0.9\textwidth]{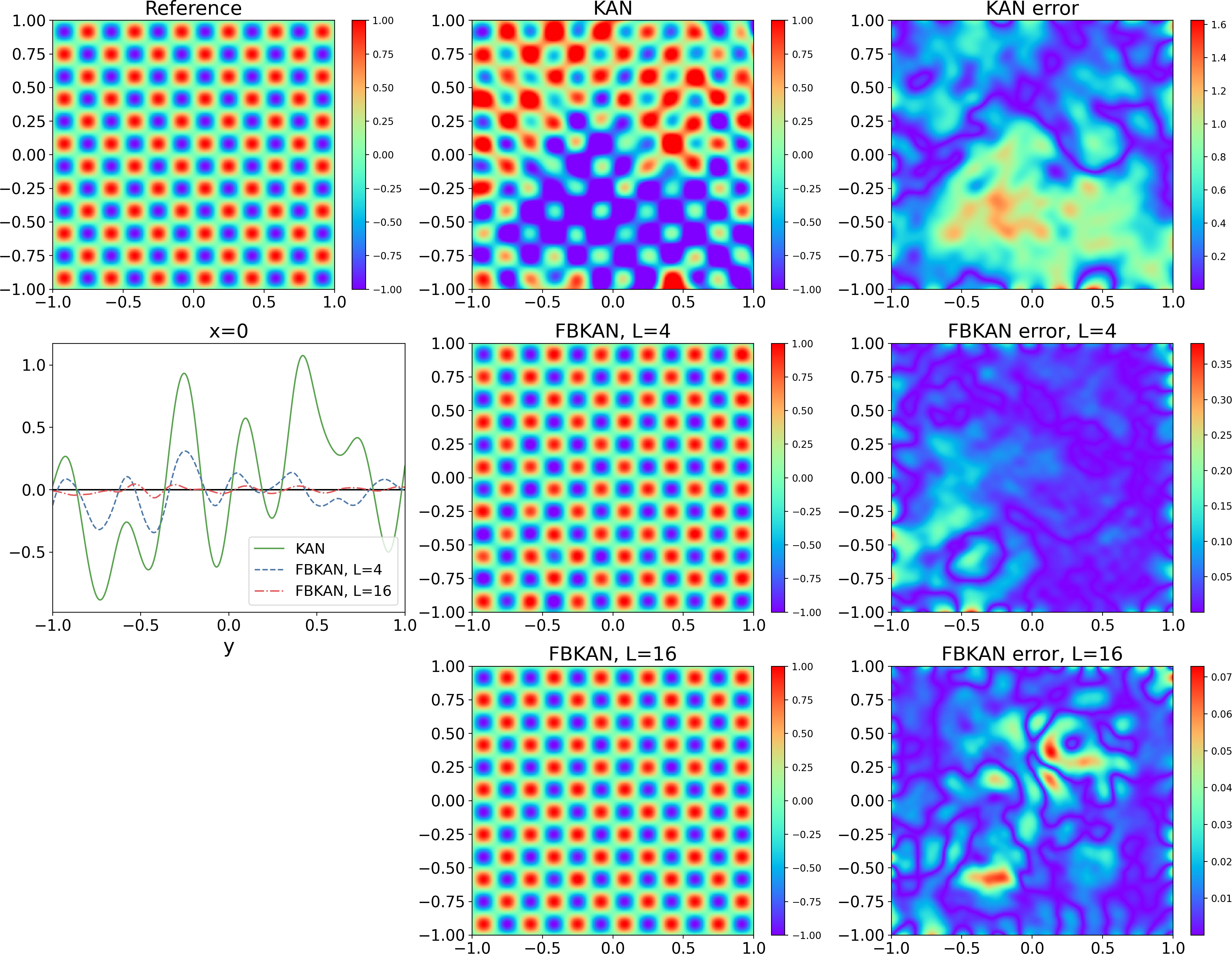}
    \caption{Results for {physics-informed test 2} with KAN-1 and FBKAN-1 and $a_1 = a_2 = 6$, with $L=4$ and $L=16$.The solution along the line $x=0$ is given in the middle left subfigure. Note that the error plots have different color scales.}
    \label{fig:Test4c}
\end{figure}

In all cases, the FBKAN outperforms the corresponding KAN {model}. The grid extension KAN cases, where we increase the grid size {during} training, comes at the expense of increasing the computational time by a factor of $1.4 -- 1.5$ in our tests.
For {$a_1 = 1$ and $a_2 = 4$}, the smaller FBKAN-3 network is sufficient to represent the solution accurately. However, for larger values of $a_1$ and $a_2$, the larger network in FBKAN-1 is necessary. The results for KAN-1 / FBKAN-1 are shown in~\cref{fig:Test4a,fig:Test4b} for {$(a_1, a_2) = (1, 4)$} and {$(a_1, a_2) = (4, 4)$}, respectively. In~\cref{fig:Test4c}, we consider $a_1=a_2=6$ with $L=4, 9$ and $16$, demonstrating the further refinement possible with additional subdomains.

\begin{table}[th]
    \centering
    \begin{tabular}{l r  r  r}
        \toprule
                      & $a_1 = 1$, $a_2 = 4$ & $a_1 = 4$, $a_2 = 4$ & $a_1 = 6$, $a_2 = 6$ \\
        \midrule
        KAN-1         & 0.0259               & 0.5465               & 1.1254               \\
        FBKAN-1, L=4  & 0.0102               & 0.0267               & 0.1151               \\
        FBKAN-1, L=9  & 0.0213               & 0.0239               & 0.0399               \\
        FBKAN-1, L=16 & 0.0037               & 0.0128               & 0.0321               \\
        KAN-2         & 0.0180               & 0.2045               & 0.5854               \\
        FBKAN-2       & 0.0112               & 0.0427               & 0.2272               \\
        KAN-3         & 0.3771               & 0.5488               & 1.2825               \\
        FBKAN-3       & 0.0214               & 0.2760               & 0.9797               \\
        \bottomrule
    \end{tabular}
    \caption{Relative $\ell_2$ errors for physics-informed {test 2}.}
    \label{tab:error-pi2}
\end{table}

\subsection{Physics-informed test 3}\label{sec:Test5}
\begin{figure}[th]
    \centering
    \includegraphics[width=0.9\textwidth]{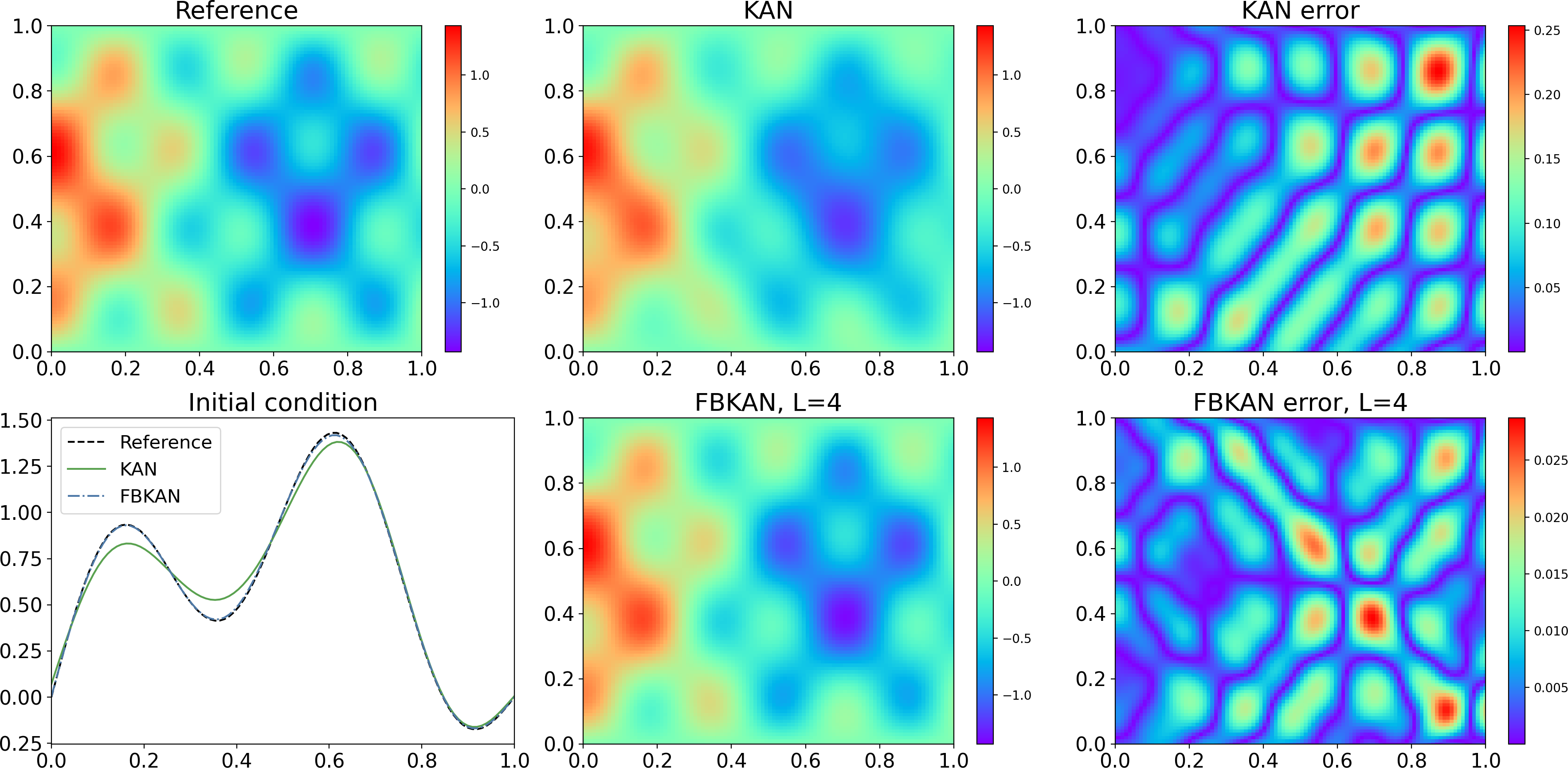}
    \caption{Results for {physics-informed test 3} with $c = \sqrt{2}$ and $L=4$. Note that the error plots have different color scales.}
    \label{fig:Test5a}
\end{figure}

\begin{figure}[th]
    \centering
    \includegraphics[width=0.9\textwidth]{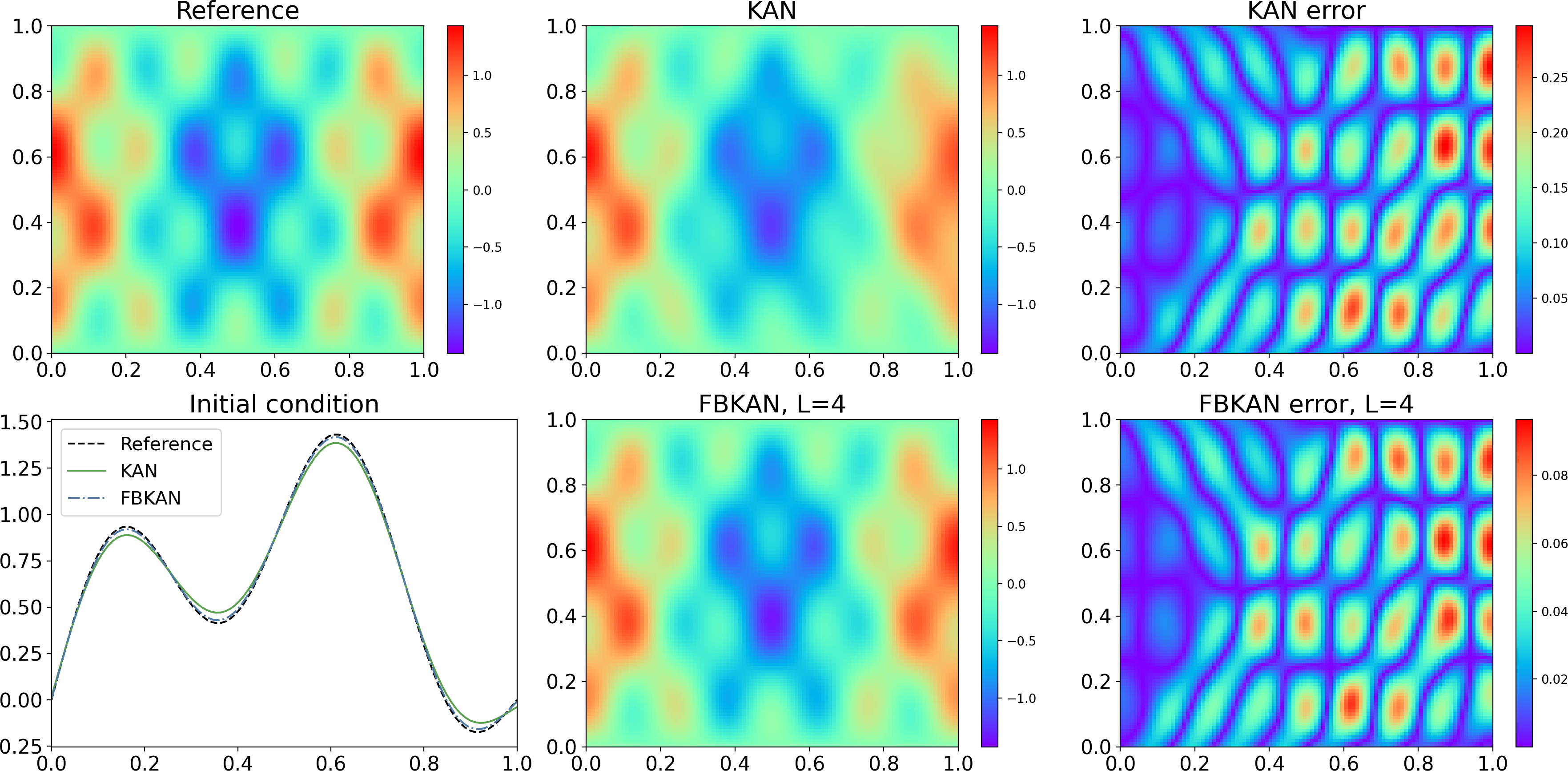}
    \caption{Results for {physics-informed test 3} with $c = 2$ and $L=4.$  Note that the error plots have different color scales.}
    \label{fig:Test5b}
\end{figure}

Finally, we consider the wave equation
\begin{align*}
    \frac{\partial^2 f}{\partial t^2} - c^2\frac{\partial^2 f}{\partial x^2} & = 0, \ (x, t) \in [0, 1]\times[0, 1],                 \\
    f(0, t)                                                                  & = 0, \  t \in [0, 1],                                 \\
    f(1, t)                                                                  & = 0, \  t \in [0, 1],                                 \\
    f(x, 0)                                                                  & = \sin( \pi x) + 0.5 \sin(4 \pi x) , \  x \in [0, 1], \\
    f_t(x, 0)                                                                & = 0, \  x \in [0, 1],
\end{align*}
which has the exact solution
$$
    f(x, t) = \sin( \pi x) \cos(c \pi t) + 0.5 \sin(4 \pi x)\cos(4 c \pi t).
$$
{For} $c = \sqrt{2}${, the KAN model} has a relative $\ell_2$ error of 0.1402, and the FBKAN with $L=4$ has a relative $\ell_2$ error of 0.0153, as illustrated in~\cref{fig:Test5a}. We then consider the {more challenging} case with $c = 2$, shown in~\cref{fig:Test5b}. The KAN has a relative $\ell_2$  error of 0.1778 and the FBKAN with $L=4$ has a relative $\ell_2$ error of 0.0587.

\section{Multilevel FBKANs}

Multilevel FBPINNs (MLFBPINNs) were introduced in~\cite{dolean_multilevel_2024} to extend FBPINNs to include multiple levels of domain decomposition simultaneously. The motivation is to allow for approximating components on various different scales. Notably, it had been observed that low-frequency components, also denoted coarse components, cannot be learned well with the one-level approach if the numbers of subdomains are being increased; this makes it necessary to add one or multiple coarser corrections; cf.~\cite{dolean_finite_2024,dolean_multilevel_2024}. In MLFBPINNs, the FBPINN architecture in~\cref{eq:FBPINN} is modified to accommodate $N$ levels of domain decomposition with each level $l$ having an overlapping domain decomposition with $J^{(l)}$ subdomains
$$
    D^{(l)} = \left\{ \Omega_j^{(l)}\right\}_{j=1}^{J^{(l)}}.
$$
We take partition of unity functions $\omega_j^{(l)}$. As in the single level case, we then require
$$
    \Omega = \cup_{j=1}^{J^{(l)}} \Omega_j^{(l)},
    \quad
    {\rm supp}\left(\omega_j^{(l)}\right) = \overline{\Omega_j^{(l)}},
    \quad \text{and} \quad
    \sum_{j=1}^{J^{(l)}} \omega_j^{(l)} \equiv 1 \text{ in } \Omega.
$$
The MLFBPINN architecture is given by
\begin{equation}
    f_\theta (\mathbf{x}) = \frac{1}{N} \sum_{l=1}^N \sum_{j=1}^{J^{(l)}} \omega_j^{(l)} (x) f_j^{(l)} (\mathbf{x}; \theta_j^{(l)}), \label{eq:MLFBPINN}
\end{equation}
where again $f_j^{(l)} (\cdot; \theta_j^{(l)})$ is the neural network with parameters $\theta_j^{(l)}$ that corresponds to the subdomain $\Omega_j^{(l)}$ in level $l$. The extension to KANs, resulting in MLFBKANs, is then straightforward:
$$
    f(\mathbf{x}) \approx \frac{1}{N} \sum_{l=1}^N \sum_{j=1}^{J^{(l)}} \omega_j^{(l)} (x) \mathcal{K}_j^{(l)}(\mathbf{x}; \theta^j), \label{eq:MLFBKAN}
$$
where $\mathcal{K}_j^{(l)}$ is the KAN corresponding to the subdomain $\Omega_j^{(l)}$ on level $l$.

In this section, we show that MLFBKANs can improve training over FBKANs for solutions with higher or multiple frequencies by considering the Helmholtz equation from~\cref{sec:Test4} and a two-dimensional multiscale Laplacian problem inspired by \cite{dolean_multilevel_2024}. In order to indicate how many subdomains on the different levels are being used we employ the notation $L = n_1, n_2, \ldots$; it means that $n_1$ subdomains are used on the first level, $n_2$ on the second level, and so on.

\subsection{Multilevel physics-informed test 1} \label{sec:ML_1}
\begin{figure}[th]
    \centering
    \includegraphics[width=0.9\textwidth]{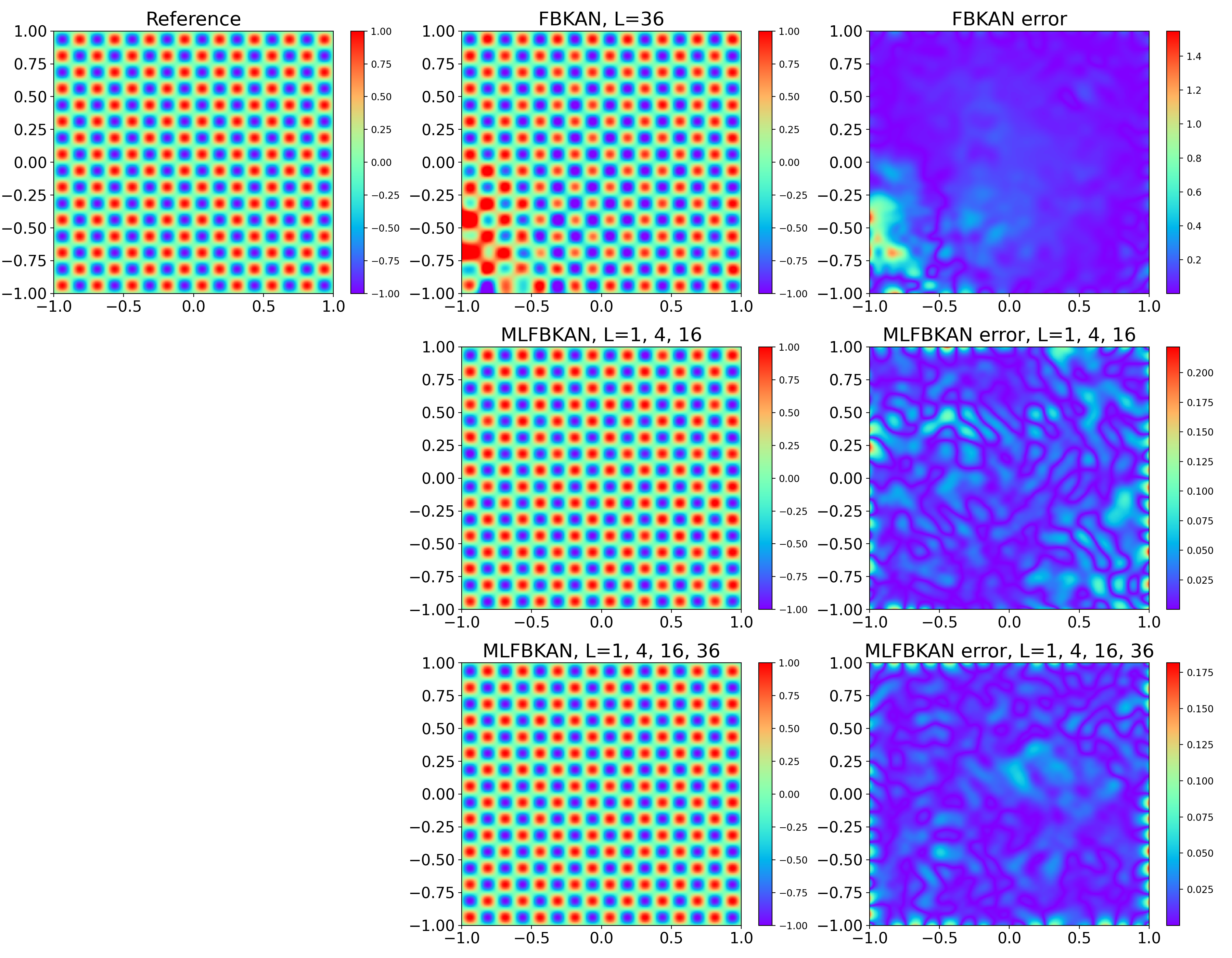}
    \caption{Results for multilevel physics-informed test 1 with $a_1 = 1, a_2 = 8$.  Note that the error plots have different color scales.}
    \label{fig:Test4a_ML}
\end{figure}

\begin{figure}[th]
    \centering
    \includegraphics[width=0.9\textwidth]{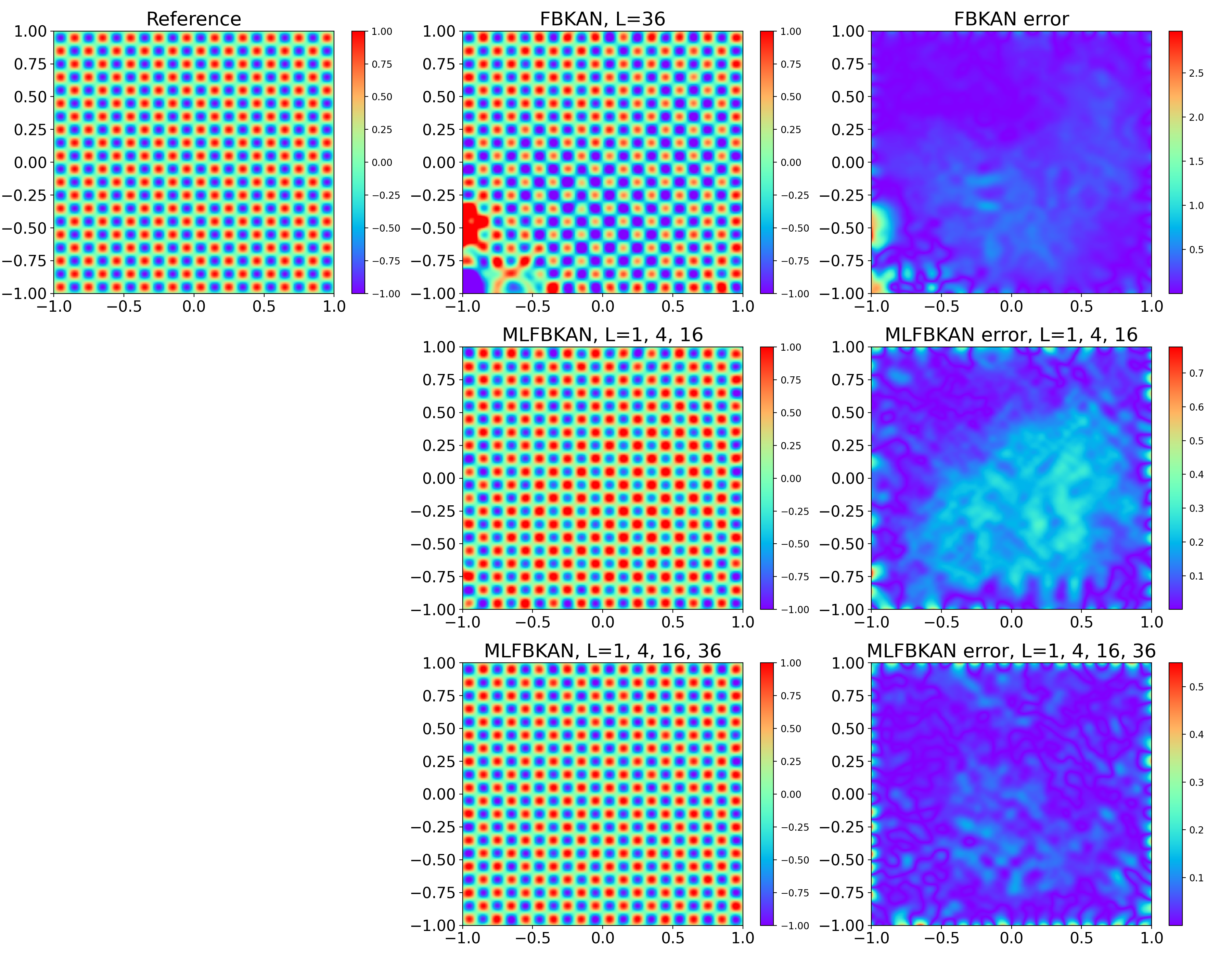}
    \caption{Results for multilevel physics-informed test 1 with $a_1 = 1, a_2 = 10$.  Note that the error plots have different color scales.}
    \label{fig:Test4b_ML}
\end{figure}

We first consider the Helmholtz equation from~\cref{sec:Test4}, choosing larger values of $a_1=a_2$ to increase the computational intensity of training the MLFBKAN. We consider $a_1 = a_2 = 8$ and $a_1 = a_2 = 10$. Error results are reported in~\cref{tab:error-ml-pi}. We compare the results across different architectures in~\cref{fig:Test4a_ML,fig:Test4b_ML}.
In both cases, the MLFBKAN with three or four levels outperforms the single level FBKAN, even with 36 subdomains. Indeed, for $a_1 = a_2 =8$, even the MLFBKAN with two levels outperforms the single level FBKANs. These results suggest that MLFBKANs can be more accurate than FBKANs for highly oscillatory problems, while reducing the total number of subdomains needed for accurate results.

\subsection{Multilevel physics-informed test 2}\label{sec:ML_2}

\begin{figure}[th]
    \centering
    \includegraphics[width=0.9\textwidth]{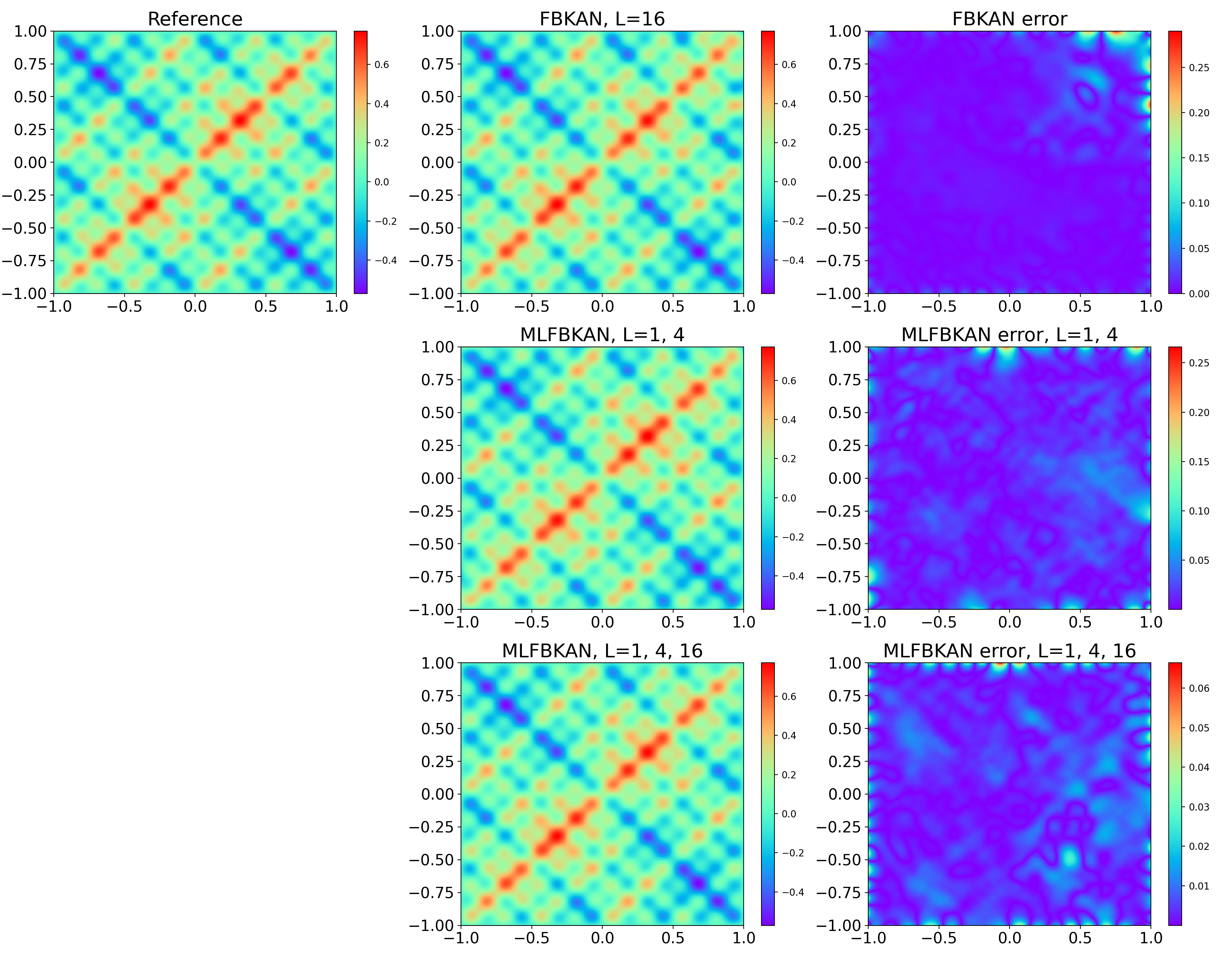}
    \caption{Results for multilevel physics-informed test 2 with $M=5$.  Note that the error plots have different color scales.}
    \label{fig:Test6a_ML}
\end{figure}

Finally, we consider a multiscale Laplacian problem also employed in \cite{dolean_multilevel_2024}.
We take
\begin{alignat*}{2}
    -\nabla^2 u(x, y) & = f(x, y),      & \quad & (x, y) \in [0, 1]\times[0, 1], \\
    u(0, y)           & = u(1, y)  = 0, &       & y \in [0, 1],                  \\
    u(x, 0)           & = u(x, 1)  = 0, &       & x \in [0, 1],
\end{alignat*}
with
$$
    f(x, y) = \frac{2}{M}\sum_{i=1}^M (2^i \pi)^2 \sin(2^i \pi x)\sin(2^i \pi y).
$$
The exact solution is then given by
$$
    u(x, y) = \frac{1}{M}\sum_{i=1}^M  \sin(2^i \pi x)\sin(2^i \pi y).
$$
From~\cref{tab:error-ml-pi}, for the case of $M= 5$, the MLFBKANs offer a significant reduction of error over the FBKANs. These predictions and errors are shown in~\cref{fig:Test6a_ML}. The errors indicate that the solution can be accurate\add{ly} captured with the finest resolution domain decomposition having 16 subdomains. 

\begin{table}[th]
    \centering
    \begin{tabular}{l | r  r | r     }
        \toprule
                                       & $a_1 = a_2 = 8$ & $a_1 = a_2 = 10$ & $M=5$   \\
        \midrule
        KAN                            & 0.89089         & 0.92729          & 1.08556 \\
        FBKAN, L=4                     & 0.91962         & 0.87100          & 0.33474 \\
        FBKAN, L=16                    & 0.28707         & 0.72251          & 0.04175 \\
        FBKAN, L=36                    & 0.33562         & 0.51766          & 0.09343 \\
        MLFBKAN, N = 2, L=1, 4         & 0.24593         & 0.87294          & 0.10461 \\
        MLFBKAN, N = 3, L=1, 4, 16     & 0.06353         & 0.27380          & 0.02926 \\
        MLFBKAN, N = 4, L=1, 4, 16, 36 & 0.04231         & 0.12012          & 0.03066 \\
        \bottomrule
    \end{tabular}
    \caption{Relative $\ell_2$ errors for the multilevel physics-informed test case 1 (middle columns) and test case 2 (right column).}
    \label{tab:error-ml-pi}
\end{table}

\section{Conclusions}
We have developed domain decomposition-based {architectures} for data-driven and physics-informed training with KANs; in accordance with {finite basis physics-informed neural networks (FBPINNs)}, we denote them as {finite basis Kolmogorov--Arnold networks (FBKANs)}.{The finite basis approach enables scalability for} complex problems{, and they} have a strong advantage over other domain decomposition-based approaches in that they do not require enforcement of transmission conditions between the subdomains via the loss function. They allow accurate training using an ensemble of small KANs combined using partition of unity functions, instead of a single large {KAN model}. We also considered MLFBKANs, following the multilevel FBPINNs~\cite{dolean_multilevel_2024} approach. Compared to the one-level approach, MLFBKANs show improvements for training with larger numbers of subdomains, which is particularly interesting for high or multi frequency problems.

Furthermore, FBKANs can be combined with existing techniques to improve the training of KANs and {physics-informed KANS (PI-KANs)}, including residual-based attention weights as introduced in~\cite{shukla2024comprehensive}, cKANs~\cite{ss2024chebyshev}, deep operator KANs~\cite{abueidda2025deepokan}, and others.
In future work, we will further examine the scalability of FBKANs and consider their application to higher dimensional problems.

\section{Code and data availability}
All code and data required to replicate the examples presented in this paper will be released upon publication. Meanwhile, we have released code and Google Colab tutorials for FBKANs in Neuromancer~\cite{Neuromancer2023} at \href{https://github.com/pnnl/neuromancer/tree/feature/fbkans/examples/KANs}{https://github.com/pnnl/neuromancer/tree/feature/fbkans/examples/KANs} for the reader to explore the ideas implemented in this work.

\section{Acknowledgments}
The KAN diagrams in~\cref{fig:graph_abs} were developed with \texttt{pykan}~\cite{liu2024kan}. This project was completed with support from the U.S. Department of Energy, Advanced Scientific Computing Research program, under the Scalable, Efficient and Accelerated Causal Reasoning Operators, Graphs and Spikes for Earth and Embedded Systems (SEA-CROGS) project (Project No. 80278) and under the Uncertainty Quantification for Multifidelity Operator Learning (MOLUcQ) project (Project No. 81739). The computational work was performed using PNNL Institutional Computing at Pacific Northwest National Laboratory. Pacific Northwest National Laboratory (PNNL) is a multi-program national laboratory operated for the U.S. Department of Energy (DOE) by Battelle Memorial Institute under Contract No. DE-AC05-76RL01830.

\bibliographystyle{unsrt}
\bibliography{references}  

\appendix
\section{Training parameters} \label{sec:training_params}
All results in this paper are implemented in \texttt{JAX}~\cite{jax2018github} using the \texttt{Jax-KAN}~\cite{Rigas_jaxKAN_A_JAX-based_2024} KAN implementation. All networks are trained with the \texttt{ADAM} optimizer. For all FBKANs we take the domain overlap $\delta=1.9$, based on the values used in \cite{dolean_multilevel_2024}.

\subsection{Data-driven test 1}
\begin{table}[h]
    \centering
    \begin{tabular}{l r r}
        \toprule
        parameter        & \cref{sec:Test1a} & \cref{sec:Test1b} \\
        \midrule
        KAN architecture & [1, 5, 1]         & [1, 5, 1]         \\
        $L$              & 4 -- 32           & 4                 \\
        $g$              & 5                 & 5                 \\
        $k$              & 3                 & 3                 \\
        learning rate    & 0.04              & 0.04              \\
        \#\,iterations   & \num{4000}        & \num{4000}        \\
        $N_{data}$       & \num{1200}        & \num{600}         \\
        \bottomrule
    \end{tabular}
    \caption{Hyperparameters used for the results {of the data-driven test 1} in~\cref{sec:Test1}.}
    \label{tab:params_Test1}
\end{table}

\subsection{Data-driven test 2}
\begin{table}[h]
    \centering
    \begin{tabular}{l r r}
        \toprule
        parameter             & KAN-1 / FBKAN-1 & KAN-2 / FBKAN-2                              \\
        \midrule
        KAN architecture      & [2, 10, 1]      & [2, 5, 1]                                    \\
        $L$                   & 4               & 4                                            \\
        $g$                   & 5               & [5, 10, 25, 30]                              \\
        $g$ schedule          & -               & [\num{0}, \num{600}, \num{1200}, \num{1800}] \\

        $k$                   & 3               & 3                                            \\
        initial learning rate & 0.02            & 0.02                                         \\
        learning rate scale   & -               & 0.8                                          \\
        \#\,iterations        & \num{2400}      & \num{2400}                                   \\
        $N_{data}$            & \num{10000}     & \num{10000}                                  \\
        \bottomrule
    \end{tabular}
    \caption{Hyperparameters used for the results {of the data-driven test 2} in~\cref{sec:Test2}. The grid ($g$) schedule denotes the iterations at which the grid is updated. The learning rate scale denotes the change to the learning rate at each grid update. KAN-1 / FBKAN-1 use fixed grids.}
    \label{tab:params_Test2}
\end{table}
\clearpage

\subsection{Physics-informed test 1}
\begin{table}[h]
    \centering
    \begin{tabular}{l  r }
        \toprule
        parameter             &                                               \\
        \midrule
        KAN architecture      & [2, 10, 1]                                    \\
        $L$                   & 4, 8                                          \\
        $g$                   & [5, 10, 15, 20]                               \\
        $g$ schedule          & [\num{0}, \num{1000}, \num{2000}, \num{3000}] \\
        $k$                   & 3                                             \\
        initial learning rate & 0.01                                          \\
        learning rate scale   & 0.8                                           \\
        \#\,iterations        & \num{4000}                                    \\
        $N_{r}$               & \num{400}                                     \\
        $N_{ic}$              & 1                                             \\
        $\lambda_{r}$         & 1/40                                          \\
        $\lambda_{ic}$        & 1                                             \\
        \bottomrule
    \end{tabular}
    \caption{Hyperparameters used for the results  {of the physics-informed test 1} in~\cref{sec:Test3}. The grid ($g$) schedule denotes the iterations at which the grid is updated. The learning rate scale denotes the change to the learning rate at each grid update. }
    \label{tab:params_Test3}
\end{table}

\subsection{Physics-informed test 2}
\begin{table}[h]
    \centering
    \begin{tabular}{l r r r}
        \toprule
        parameter             & KAN-1 / FBKAN-1                        & KAN-2 / FBKAN-2             & KAN-3 / FBKAN-3 \\
        \midrule
        KAN architecture      & [2, 10, 1]                             & [2, 10, 1]                  & [2, 5, 1]       \\
        $L$                   & 4 -- 16                                & 4                           & 4               \\
        $g$                   & 5                                      & [5, 10, 15]                 & 5               \\
        $g$ schedule          & -                                      & [0, \num{3000}, \num{6000}] & -               \\
        $k$                   & 5                                      & 3                           & 5               \\
        initial learning rate & 0.005                                  & 0.005                       & 0.005           \\
        learning rate scale   & -                                      & 0.8                         & -               \\
        \#\,iterations        & \makecell{$a_1=1, a_2 =4$: \num{10000}                                                 \\ $a_1=4, a_2 =4$: \num{10000} \\ $a_1=6, a_2 =6$: \num{30000}} & \num{10000} & \num{10000} \\
        $N_{r}$               & 800                                    & 800                         & 800             \\
        $N_{bc}$              & 400                                    & 400                         & 400             \\
        $\lambda_{r}$         & 0.01                                   & 0.01                        & 0.01            \\
        $\lambda_{bc}$        & 1                                      & 1                           & 1               \\
        \bottomrule
    \end{tabular}
    \caption{Hyperparameters used for the results  {of the physics-informed test 2} in~\cref{sec:Test4}. The grid ($g$) schedule denotes the iterations at which the grid is updated. The learning rate scale denotes the change to the learning rate at each grid update. }
    \label{tab:params_Test4}
\end{table}
\clearpage

\subsection{Physics-informed test 3}
\begin{table}[h]
    \centering
    \begin{tabular}{l r r}
        \toprule
        parameter             & $ c =  \sqrt{2}$ & $c = 2$        \\
        \midrule
        KAN architecture      & [2, 10, 1]       & [2, 10, 10, 1] \\
        $L$                   & 4                & 4              \\
        $g$                   & 10               & 10             \\
        $k$                   & 5                & 5              \\
        initial learning rate & 0.001            & 0.0005         \\
        \#\,iterations        & \num{60000}      & \num{120000}   \\
        $N_{r}$               & \num{1000}       & \num{1200}     \\
        $N_{ic}$              & 100              & 100            \\
        $N_{bc}$              & 200              & 200            \\
        $\lambda_{r}$         & 0.01             & 0.01           \\
        $\lambda_{ic}$        & 1                & 1              \\
        $\lambda_{bc}$        & 1                & 1              \\
        \bottomrule
    \end{tabular}
    \caption{Hyperparameters used for the results  {of the physics-informed test 3} in~\cref{sec:Test5}. }
    \label{tab:params_Test5}
\end{table}

\subsection{Multilevel physics-informed test 1}
\begin{table}[h]
    \centering
    \begin{tabular}{l r}
        \toprule
        parameter             &             \\
        \midrule
        KAN architecture      & [2, 10, 1]  \\
        $L$                   & 4 -- 36     \\
        $N$                   & 1 -- 4      \\
        $g$                   & 5           \\
        $g$ schedule          & -           \\
        $k$                   & 5           \\
        initial learning rate & 0.005       \\
        learning rate scale   & -           \\
        \#\,iterations        & \num{30000} \\
        $N_{r}$               & 800         \\
        $N_{bc}$              & 400         \\
        $\lambda_{r}$         & 0.01        \\
        $\lambda_{bc}$        & 1           \\
        \bottomrule
    \end{tabular}
    \caption{Hyperparameters used for the results  {of the multilevel physics-informed test 1} in~\cref{sec:ML_1}. The grid ($g$) schedule denotes the iterations at which the grid is updated. The learning rate scale denotes the change to the learning rate at each grid update. }
    \label{tab:params_Test5b}
\end{table}
\clearpage

\subsection{Multilevel physics-informed test 2}
\begin{table}[h]
    \centering
    \begin{tabular}{l r}
        \toprule
        parameter             &             \\
        \midrule
        KAN architecture      & [2, 10, 1]  \\
        $L$                   & 4 -- 36     \\
        $N$                   & 1 -- 4      \\
        $g$                   & 5           \\
        $g$ schedule          & -           \\
        $k$                   & 5           \\
        initial learning rate & 0.005       \\
        learning rate scale   & -           \\
        \#\,iterations        & \num{30000} \\
        $N_{r}$               & 800         \\
        $N_{bc}$              & 400         \\
        $\lambda_{r}$         & 0.001       \\
        $\lambda_{bc}$        & 1           \\
        \bottomrule
    \end{tabular}
    \caption{Hyperparameters used for the results  {of the multilevel physics-informed test 2} in~\cref{sec:ML_2}. The grid ($g$) schedule denotes the iterations at which the grid is updated. The learning rate scale denotes the change to the learning rate at each grid update. }
    \label{tab:params_Test6}
\end{table}
\afterpage{\clearpage}

\end{document}